\documentclass[a4paper]{article}

\usepackage[utf8]{inputenc}
\usepackage[T1]{fontenc}
\usepackage{hyperref,url}

\usepackage{booktabs}
\usepackage{nicefrac}
\usepackage{microtype}
\usepackage{pifont}

\usepackage{graphicx}
\graphicspath{{./}{./figures/}}
\usepackage{subcaption}

\usepackage{tabularx,ragged2e}
\newcolumntype{L}{>{\RaggedRight}X}

\usepackage{mathtools,empheq,amsfonts,amsthm,amsmath,amssymb,bm,mathrsfs}

\usepackage[title]{appendix}
\usepackage{xcolor}
\usepackage{textcomp}
\usepackage{manyfoot}
\usepackage{booktabs}
\usepackage{listings}

\usepackage{anyfontsize}
\usepackage{cleveref}
\usepackage{soul}
\AtBeginEnvironment{appendices}{\crefalias{section}{appendix}}

\newcommand{\rev}[1]{{#1}}

\newcommand{\Jac}{\mathrm{J}}
\newcommand{\fim}{\mathcal{I}}
\newcommand{\ofim}{\mathfrak{J}}
\newcommand{\expect}{\mathbb{E}}
\newcommand{\trace}{\mathrm{tr}}

\newtheorem{theorem}{Theorem}
\newtheorem{lemma}[theorem]{Lemma}

\newtheorem{proposition}[theorem]{Proposition}
\newtheorem{remark}{Remark}\newtheorem{definition}{Definition}

\renewcommand{\top}{\intercal}

\newcommand{\de}{\mathrm{d}}
\newcommand{\dx}{\mathrm{d}\bm{x}}
\newcommand{\dtheta}{\mathrm{d}\bm{\theta}}
\newcommand{\deta}{\mathrm{d}\bm{\eta}}
\newcommand{\dxi}{\mathrm{d}\bm{\xi}}
\newcommand{\dzeta}{\mathrm{d}\bm{\zeta}}
\newcommand{\deu}{\mathrm{d}_{\mathrm{E}}}
\newcommand{\cov}[1]{\mathrm{cov}(#1)}
\newcommand{\mean}[1]{\mu(#1)}
\newcommand{\diag}[1]{\mathrm{diag}\left({#1}\right)}
\newcommand{\rank}[1]{\mathrm{rank}\left({#1}\right)}

\renewcommand{\Re}{\mathbb{R}}

\newcommand{\calG}{\mathcal{G}}
\newcommand{\calM}{\mathcal{M}}
\newcommand{\tildeM}{\widetilde{\mathcal{M}}}

\newcommand{\calT}{\mathcal{T}}
\newcommand{\calO}{\mathcal{O}}
\newcommand{\calX}{\mathcal{X}}

\newcommand{\calZ}{\mathcal{Z}}
\newcommand{\calTM}{\mathcal{TM}}
\newcommand{\rad}{\mathrm{Rad}(\mathcal{TM})}

\newcommand{\scr}{\mathcal{S}(\mathcal{TM})}

\newcommand{\onehot}{\mathrm{OneHot}}
\newcommand{\softmax}{\mathrm{SoftMax}}

\newcommand{\defeq}{\vcentcolon=}

\usepackage{xspace}
\makeatletter
\DeclareRobustCommand\onedot{\futurelet\@let@token\bmv@onedotaux}
\def\bmv@onedotaux{\ifx\@let@token.\else.\null\fi\xspace}
\def\eg{\emph{e.g}\onedot} 
\def\ie{\emph{i.e}\onedot} 
 
 \def\vs{\emph{vs}\onedot}
\def\wrt{w.r.t\onedot} 

\def\iid{i.i.d\onedot}

\makeatother

\usepackage[a4paper]{geometry}
\usepackage{orcidlink}

\usepackage[square,sort,comma,numbers]{natbib}
\definecolor{mydarkblue}{rgb}{0,0.08,0.45}
\definecolor{highlightyellow}{RGB}{255, 250, 205}
\hypersetup{ pdftitle={},
    pdfauthor={},
    pdfsubject={},
    pdfkeywords={},
    pdfborder=0 0 0,
    pdfpagemode=UseNone,
    colorlinks=true,
    linkcolor={red!80!white},
    citecolor=mydarkblue,
    filecolor=mydarkblue,
    urlcolor=mydarkblue,
    pdfview=FitH
}

\title{A Geometric Modeling of Occam's Razor in Deep Learning}
\author{Ke Sun~\orcidlink{0000-0001-6263-7355}\\
CSIRO's Data61, Australia\\
Ke.Sun@data61.csiro.au, sunk@ieee.org
\and
Frank Nielsen~\orcidlink{0000-0001-5728-0726}\\
Sony Computer Science Laboratories Inc. (Sony CSL)\\
Tokyo, Japan\\
Frank.Nielsen@acm.org
}

\date{Version: June 2025}

\makeatletter
\begin{document}

\maketitle

\begingroup
\renewcommand\thefootnote{}\footnote{Cite as:~\colorbox{highlightyellow}{\parbox{.85\linewidth}{Sun, Ke and Nielsen, Frank, \textbf{A Geometric Modeling of Occam's Razor in Deep Learning}, \textit{Information Geometry},
Special Issue: Half a Century of Information Geometry, Part 2, 2025. DOI: \url{https://doi.org/10.1007/s41884-025-00167-2}}}}\addtocounter{footnote}{-1}
\footnote{This work first appeared under the former title ``Lightlike Neuromanifolds, Occam's Razor and Deep Learning'' in 2019.}
\addtocounter{footnote}{-1}
\endgroup

\begin{abstract}Why do deep neural networks (DNNs) benefit from very high dimensional parameter spaces?
Their huge parameter complexities \vs stunning performance in practice is all the more intriguing and not explainable using the standard theory of model selection for regular models.
In this work, we propose a geometrically flavored information-theoretic approach to study this phenomenon.
\rev{With the belief that simplicity is linked to better generalization, as grounded in the theory of minimum description length, the objective of our analysis is to examine and bound the complexity of DNNs.}
We introduce the locally varying dimensionality of the parameter space of neural network models
by considering the number of significant dimensions of the Fisher information matrix, and model the parameter space as a manifold using the framework of singular semi-Riemannian geometry.
We derive model complexity measures which yield short description lengths
for deep neural network models based on their singularity analysis thus explaining the good performance of DNNs despite their large number of parameters.
 \end{abstract}
\noindent\textbf{Keywords:~}{Information geometry, Deep learning, Minimum Description Length, Fisher information, Stochastic complexity}

\section{Introduction}\label{sec:intro}

Deep neural networks (DNNs) are usually large models in terms of storage costs. In
the classical model selection theory, such models are not favored as compared to
simple models with the same training performance. For example, if one applies
the Bayesian information criterion (BIC)~\citep{sETD} to DNN, a shallow neural
network (NN) will be preferred over a deep NN due to the penalty term with respect to (\wrt) the complexity.  A basic principle in
science is the Occam\footnote{William of Ockham (ca. 1287 --- ca. 1347), a monk
(friar) and philosopher.}'s Razor, which favors simple models over complex ones
that accomplish the same task. This raises the fundamental question of
\emph{how to measure the simplicity or the complexity of a model}.

Formally, the preference of simple models has been studied in the area of
minimum description length (MDL)~\cite{rMBS,rFIA,pTMDL}, also known in another
thread of research as the minimum message length (MML)~\cite{wbAIM}.
\rev{By the theory of MDL~\cite{pTMDL}, statistical models that can most concisely communicate the observed data are favored and expected to generalize better~\cite{grMDL,MDLbarroncover,MDLzhang,pmlr-v98-grunwald19a,PACMDLbounds}. This is intuitive, as complex models often lead to overfit.}

Consider a parametric family of distributions
$\calM=\{p(\bm{x}\mid\bm\theta)\}$ with $\bm{\theta}\in\Theta\subset\Re^D$.
The distributions are mutually absolutely continuous,
which guarantees all densities to have the same support.
Otherwise, many problems of non-regularity will arise as described
by \cite{InfiniteFIM2011,InsufficiencyFisher2013}.
The Fisher information matrix (FIM) $\fim(\bm\theta)$ is
a $D\times{D}$ positive semi-definite (psd) matrix:
$\fim(\bm\theta)\succeq0$.
The model is called \emph{regular} if it is
 (i) identifiable~\citep{calin2014geometric} with (ii) a non-degenerate and finite Fisher information
matrix (i.e., $\fim(\bm\theta)\succ 0$).

In a Bayesian setting, the description length of
a set of $N$ \iid observations $\bm{X}=\{\bm{x}_i\}_{i=1}^N\subset\calX$ \wrt
$\calM$ can be defined as the number of \emph{nats} with the coding scheme of a
parametric model $p(\bm{x}\,\vert\,\bm{\theta})$ and a prior $p(\bm\theta)$.
The code length of any $\bm{x}_i$ is given by the cross entropy
between the empirical distribution $\delta_i(\bm{x})= \delta(\bm{x}-\bm{x}_i)$,
where $\delta(\cdot)$ denotes the Dirac's delta function, and
$p(\bm{x})= \int p(\bm{x}\mid\bm\theta) p(\bm\theta)\, \dtheta$.
Therefore, the description length of $\bm{X}$ is
\begin{equation}\label{eq:bayes}
    - \log p(\bm{X})
    = \sum_{i=1}^N h^\times(\delta_i : p)
    = -\sum_{i=1}^N \log \int p(\bm{x}_i \mid \bm\theta)\, p(\bm\theta)\,\dtheta,
\end{equation}
where $h^\times(p:q) \defeq -\int p(\bm{x})\log q(\bm{x})\dx$
denotes the cross entropy between $p(\bm{x})$ and $q(\bm{x})$, and $\log$ denotes natural logarithm throughout the paper. The code length means the cumulative loss of the Bayesian mixture model $p(\bm{x})$ \wrt the observations $\bm{X}$.
\rev{\Cref{eq:bayes} corresponds to the Bayesian universal code.
In MDL, the optimal code in terms of the minimax strategy~\cite{shtarkov87} is given by the normalized maximum likelihood (NML) code. With a suitable choice of the prior, the Bayesian universal code and the NML code asymptotically coincide~\cite{pTMDL} with O(1) difference.}

By using Jeffreys\footnote{Sir Harold Jeffreys (1891--1989), a British statistician.}'
non-informative prior~\citep{aIGA} as $p(\bm\theta)$, the MDL in \cref{eq:bayes}
can be approximated (see \citep{rMBS,rFIA,bMDL}) as
\begin{equation}\label{eq:mdl}
\chi = \underbrace{- \log p(\bm{X} \mid \hat{\bm\theta})}_{\mbox{fitness}}
+ \overbrace{\underbrace{\frac{D}{2}\log\frac{N}{2\pi}}_{\mbox{penalize high dof}}
+ \underbrace{\log \left( \int \sqrt{\vert\fim(\bm\theta)\vert}\;d\bm{\theta} \right)}_{\mbox{model capacity}}}^{\mbox{geometric complexity}},
\end{equation}
where $\hat{\bm\theta}\in\Theta$ is the maximum likelihood estimation (MLE),
or the projection~\citep{aIGA} of $\bm{X}$ onto the model,
$D=\dim(\Theta)$ is the model size, $N$ is the number of observations,
and $\vert\cdot\vert$ denotes the matrix determinant.
In this paper, the symbols $\chi$ and $\calO$ and the term ``razor'' all refer
to the same concept, that is the description length of the data $\bm{X}$ by
the model $\calM$. The smaller those quantities, the better.

The first term in~\cref{eq:mdl} is the fitness of the model to the observed data.
The second and the third terms measure the \emph{geometric complexity}~\citep{mbCPD}
and make $\chi$ favor simple models.
The second $O(\log{N})$ term only depends on the number of parameters $D$
and the number of observations $N$. It penalizes large models with a high degree of freedom (dof).
The third $O(1)$ term is independent to the observed data and measures the model capacity,
or the total ``number'' of distinguishable distributions~\citep{mbCPD} in the model.

Unfortunately, this razor $\chi$ in \cref{eq:mdl} does not fit straightforwardly
into DNNs, which are high-dimensional \emph{singular} models. The FIM $\fim(\bm\theta)$
is a large singular matrix (not full rank) and the last term may be difficult to evaluate.
Based on the second term on the right-hand-side (RHS), a DNN can
have very high complexity and therefore is less favored against a shallow
network. This contradicts the good generalization of DNNs as compared to shallow NNs.
These issues call for a new analysis of the MDL in the DNN setting.

Towards this direction, we made the following contributions in this paper:
\begin{itemize}
    \item[--] New concepts and methodologies from singular semi-Riemannian
        geometry~\citep{kSSR} to analyze the space of neural networks;
    \item[--] A definition of the local dimensionality in this space, that is the amount of
        non-singularity, with bounding analysis;
    \item[--] \rev{A connection between $f$-mean and DNN model complexity with related bounds;}
    \item[--] A new MDL formulation, which explains how the singularities
        contribute to the ``negative complexity'' of DNNs: That is, the model turns
        simpler as the number of parameters grows.
\end{itemize}

The rest of this paper is organized as follows.
\Cref{sec:ssrg} reviews singularities in information geometry.
In the setting of a DNN,
\cref{sec:lightdl} introduces its singular parameter manifold.
\Cref{sec:localdim} bounds the number of singular dimensions of the parameter manifold of the DNN.
\Cref{sec:general,sec:fmean,sec:gauss,sec:jeffreys} derive
our MDL criterion based on two different priors,
and discuss how model complexity is affected by the singular geometry.
We discuss related work in \cref{sec:related} and conclude in \cref{sec:con}.
\rev{Proofs and related derivations of our main results are provided in the appendix.}

\section{Lightlike Statistical Manifold}\label{sec:ssrg}

In this paper, bold capital letters like $\bm{A}$ denote matrices,
bold small letters like $\bm{a}$ denote vectors, normal capital/small letters
like $A$/$a$ and Greek letters like $\alpha$ denote scalars,
and calligraphy letters like $\calM$ denote manifolds (with exceptions).
We use $\bm{X}$, $\bm{Y}$ and $\bm{Z}$ to denote a collection of $N$ random observations and use $\bm{x}$, $y$ and $\bm{z}$ to denote one single observation.

The term ``statistical manifold'' refers to
$\calM=\left\{p(\bm{x}\,\vert\,\bm{\theta})\right\}$,
where each point of $\calM$ corresponds to a probability distribution $p(\bm{x}\,\vert\,\bm{\theta})$\footnote{To be more precise, a statistical manifold~\citep{Lauritzen-1987} is a structure $(\nabla,g,C)$ on a smooth manifold $\calM$,
where $g$ is a metric tensor, $\nabla$ a torsion-free affine connection, and $C$ is a symmetric covariant tensor of order 3.}.
The discipline of information geometry~\citep{aIGA} studies such a space in
the Riemannian and more generally differential geometry framework.
Hotelling~\citep{Hotelling-1930} and independently
Rao~\citep{Rao-1945,Rao-reprint-1992} proposed to endow a
parametric space of statistical models
with the Fisher information matrix as a Riemannian metric:
\begin{equation}\label{eq:fim}
\fim(\bm\theta)
\defeq
\expect_p \left( \frac{\partial\log p(\bm{x}\,\vert\,\bm{\theta})}{\partial\bm\theta}
\frac{\partial\log p(\bm{x}\,\vert\,\bm\theta)}{\partial\bm\theta^\top} \right),
\end{equation}
where $\expect_p$ denotes the expectation \wrt $p(\bm{x}\,\vert\,\bm\theta)$.
The corresponding infinitesimal squared length element $\mathrm{d}s^2 =
\trace(\fim(\bm\theta) d\bm\theta d\bm\theta^\top)
=
\langle d\bm\theta,d\bm\theta \rangle_{\fim(\bm\theta)}=d\bm\theta^\top \fim(\bm\theta) d\bm\theta$,
where $\trace(\cdot)$ means the matrix trace\footnote{Using the cyclic property of the matrix trace, we
have $\mathrm{d}s^2=\trace(\fim(\bm\theta)\dtheta\dtheta^\top)=\dtheta^\top\fim(\bm\theta)\dtheta$.},
is independent of the underlying parameterization of the population space.

Amari further developed this approach by revealing the dualistic
structure of statistical manifolds which extends the Riemannian framework~\citep{aIGA,nkADG}.
The MDL criterion arising from the geometry of Bayesian inference
with Jeffreys' prior for regular models is detailed in~\citep{bMDL}.
In information geometry, the regular assumption is
(1) an open connected parameter space
in some Euclidean space; and
(2) the FIM exists and is non-singular.
However, in general, the FIM is only positive semi-definite and thus for
non-regular models like neuromanifolds~\citep{aIGA} or Gaussian
mixture models~\citep{wAGA}, the manifold is not Riemannian but
\emph{singular semi-Riemannian}~\citep{kSSR,dbLSO}.
In the machine learning community, singularities have often been dealt with
as \emph{a minor issue}:
For example, the natural gradient has been generalized based on the
Moore-Penrose inverse of $\fim(\bm\theta)$~\citep{tGAG} to avoid potential
non-invertible FIMs.
Watanabe~\cite{wAGA} addressed the fact that most usual learning machines are
singular in his singular learning theory which
relies on algebraic geometry.
Nakajima and Ohmoto~\cite{nakajima2021dually} discussed dually flat structures for singular models.

Recently, preliminary efforts~\citep{btGOL,jsOTG} tackle singularity
at the core, mostly from a mathematical standpoint.
For example, Jain et al.~\citep{jsOTG} studied the Ricci curvature tensor
of such manifolds.
These mathematical notions are used in the community of differential
geometry or general relativity
but have not yet been ported to the machine learning community.

\begin{figure}[tb]
\centering
\includegraphics[width=.65\textwidth]{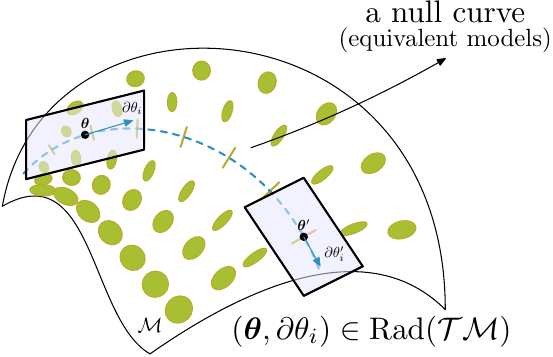}
\caption{A toy lightlike manifold $\calM$ with a null curve. The ellipses are
    Tissot's indicatrices, showing how circles of infinitesimal radius
    are distorted by the lightlike geometry on $\calM$. On the null curve, the
    FIM is degenerate so that
    $\langle\partial\theta_i,\partial\theta_i\rangle_{\fim}=0$.
    Therefore the local dynamic $\partial\theta_i$ (tangent vector of the null
    curve) has zero length, meaning that it does not change the model.
    The radical distribution $\rad$ is formed by the null curve and
    its tangent vectors.
    \rev{In the context of DNNs, such dynamics refer to small changes of NN weights/biases that do not alter the global model.}
    }\label{fig:toy}
\end{figure}

Following these efforts, we first introduce informally some basic
concepts from a machine learning perspective
to define the differential geometry of non-regular statistical manifolds.
The \emph{tangent space} $\calT_{\bm\theta}(\calM)$ is a
$D$-dimensional ($D=\dim(\calM)$) real vector space,
that is the local linear approximation
of the manifold $\calM$ at the point $\bm\theta\in\calM$,
equipped with the inner product induced
by $\fim(\bm\theta)$. The \emph{tangent bundle}
$\calTM \defeq \{({\bm\theta},\bm v), \bm\theta\in\calM, \bm v\in \calT_{\bm\theta}\}$
is the $2D$-dimensional manifold obtained by combining
all tangent spaces for all $\bm\theta\in\calM$.
A \emph{vector field} is a smooth mapping from $\calM$ to $\calTM$ such
that each point $\bm\theta\in\calM$ is attached a tangent vector originating from itself.
Vector fields are cross-sections of the tangent bundle.
In a local coordinate chart $\bm\theta$, the vector fields along the frame are denoted as $\partial\theta_i$.
A \emph{distribution} (not to be confused with probability distributions
which are points on $\calM$)
means a vector subspace of the tangent bundle spanned by several independent
vector fields, such that each point $\bm\theta\in\calM$
is associated with a subspace of
$\calT_{\bm\theta}(\calM)$ and those subspaces vary smoothly with $\bm\theta$.
Its dimensionality is defined by the dimensionality of the subspace,
\ie, the number of vector fields that span the distribution.

In a \emph{lightlike} manifold~\citep{kSSR,dbLSO} $\calM$, $\fim(\bm\theta)$ can be degenerate.
The tangent space $\calT_{\bm\theta}(\calM)$ is a vector space with a kernel subspace, \ie, a nullspace.
A null vector field is formed by null vectors, whose lengths
measured according to the Fisher metric tensor are all zero.
The \emph{radical~\footnote{\emph{Radical} stems from Latin and means root.} distribution} $\rad$ is the distribution
spanned by the null vector fields. Locally at $\bm\theta\in\calM$,
the tangent vectors in $\calT_{\bm\theta}(\calM)$ which span the
kernel of $\fim(\bm\theta)$ are denoted as
$\mathrm{Rad}_{\bm\theta}(\calTM)$. In a local coordinate chart,
$\rad$ is well defined if these
$\mathrm{Rad}_{\bm\theta}(\calTM)$ form a valid distribution.
We write $\calTM=\rad\oplus\scr$, where
`$\oplus$'' is the direct sum, and the \emph{screen distribution}
$\scr$ is complementary to the radical distribution $\rad$ and has
a non-degenerate induced metric.  See \cref{fig:toy} for an illustration of the
concept of radical distribution.

We can find a local coordinate frame (a frame is an ordered basis)
$\left(\theta_1,\cdots,\theta_d,\theta_{d+1},\cdots,\theta_D\right)$, where
the first $d$ dimensions $\bm\theta^{s}=(\theta_1,\cdots,\theta_d)$ correspond
to the screen distribution, and the remaining $\bar{d}\defeq{}D-d$ dimensions
$\bm\theta^{r}=(\theta_{d+1},\cdots,\theta_D)$ correspond to the radical
distribution.
The local inner product $\langle\cdot,\cdot\rangle_{\fim}$ satisfies
\begin{align*}
\langle \partial\theta_i,\partial \theta_j\rangle_{\fim} &= \delta_{ij},
\quad(\forall\ {1}\le{i,j}\le{d})\\
\langle \partial\theta_i,\partial \theta_k\rangle_{\fim} &= 0,
\quad(\forall\ {d+1}\le{i}\le{}D,\;1\le{k}\le{D})
\end{align*}
where $\delta_{ij}=1$ if and only if (iff) $i=j$ and $\delta_{ij}=0$, otherwise.
Unfortunately, this frame is not unique~\citep{dARO}. We will abuse $\fim$ to denote
both the FIM of $\bm\theta$ and the FIM of $\bm\theta^{s}$. One has to
remember that $\fim(\bm\theta)\succeq0$, while $\fim(\bm\theta^s)\succ0$ is a
proper Riemannian metric.  Hence, both $\fim^{-1}(\bm\theta^s)$ and
$\log\vert\fim(\bm\theta^s)\vert$ are well-defined.

\begin{remark}
Notice that the Fisher information matrix is \emph{covariant} under
reparameterization.
That is, let $\bm\theta(\bm\lambda)$ be an invertible smooth reparameterization of
$\bm\lambda$.
Then the FIM rewrites in the $\theta$-parameterization as:
\begin{equation}\label{eq:HFIM}
\fim(\bm\theta)
= \Jac_{\bm\theta\to\bm\lambda}^\top
\fim(\bm\lambda(\bm\theta))
\Jac_{\bm\theta\to\bm\lambda},
\end{equation}
where $\Jac_{\bm\theta\to\bm\lambda}$ is the full rank Jacobian matrix.

The natural gradient flows (vector fields on $\mathcal{M}$)
with respect to $\bm\lambda$ and $\bm\theta$ coincide but
not the natural gradient descent methods (learning paths
that consist of sequences of points on $\mathcal{M}$)
because of the non-zero learning step sizes.

Furthermore, the ranks of $\fim(\bm\theta)$ and $\fim(\bm\lambda)$
as well as the dimensions of the screen and radical distributions coincide.
Hence, the notion of singularities is intrinsic and independent of the smooth reparameterization.
\end{remark}

\section{Lightlike Neuromanifold}\label{sec:lightdl}

This section instantiates the concepts in the previous \cref{sec:ssrg} in terms of a simple DNN
predictive model. The random variable $\bm{x}=(\bm{z},y)$ of interest
consists of two components: $\bm{z}$, referred to as the ``input'',
and $y$, referred to as the ``target''.
By assumption, their joint probability distribution is specified by
\begin{equation*}
\log p(\bm{x} \mid \bm\psi, \bm\theta)
=
\log p(\bm{z} \mid \bm\psi)
+
\log p(y \mid \bm{z},\,\bm\theta),
\end{equation*}
where $p(\bm{z} \mid \bm\psi)$ is a generative model of $\bm{z}$ which is
parameterized by $\bm\psi$, $p(y \mid \bm{z},\,\bm\theta)$ is a predictive
DNN, and $\bm\theta$ consists of all neural network parameters.

Our main subject is the latter predictive model $p(y \mid \bm{z},\,\bm\theta)$ and its parameter manifold
$\calM_{\bm\theta}$.
Here, we need the generative model $p(\bm{z} \mid \bm\psi)$ for the purpose of
discussing how the geometry of $\calM_{\bm\theta}$ is affected by the
choice of $p(\bm{z} \mid \bm\psi)$ and can be studied independent of the
parameter space of $p(\bm{z} \mid \bm\psi)$, which we denote as $\calM_{\bm\psi}$.
In the end, our results do not depend on the specific form of $p(\bm{z})$ or whether it is parametric.

For $p(y \mid \bm{z},\,\bm\theta)$, we consider a deep feed-forward network with $L$ layers,
uniform width $M$ except the last layer which has $m$ output units ($m<M$),
input $\bm{z}\in\calZ$ with $\dim(\calZ)=M$,
pre-activations $\bm{h}^l$ of size $M$ (except that in the last layer, $\bm{h}^L$ has $m$ elements),
post-activations $\bm{z}^l$ of size $M$,
weight matrices $\bm{W}^l$
and bias vectors $\bm{b}^l$ ($1\le{l}\le{L}$).
The layers are given by
\begin{equation}\label{eq:nn}
\begin{aligned}
    \bm{z}^l &= \phi(\bm{h}^l),\\
    \bm{h}^l &= \bm{W}^l \bm{z}^{l-1} + \bm{b}^{l},\\
    \bm{z}^0 &= \bm{z},
\end{aligned}
\end{equation}
where $\phi$ is an element-wise nonlinear activation function such as ReLU~\citep{gbDSR}.

Without loss of generality, we assume multinomial\footnote{In fact, a generalization of the Bernoulli distribution with integer $k\geq 2$ mutually exclusive events, called informally a multinoulli distribution since it is a multinomial distribution with a single trial.} output units
and the DNN output~\cite{goodfellow2016deep}
\begin{equation*}
y \sim \mathrm{Multinomial}\left( \softmax(\bm{h}^L) \right)
\end{equation*}
is a random label in the set $\{1,\cdots,m\}$, where
\begin{equation*}
\softmax(\bm{t})\defeq\frac{1}{\sum_{i=1}^m\exp(t_i)}\left(\exp(t_1),\exp(t_2),\cdots,\exp(t_m)\right)
\end{equation*}
denotes the softmax function. $\softmax(\bm{h}^L)$ is a random point in
$\Delta^m$, the $(m-1)$ dimensional statistical simplex.
Therefore,
$p(y=k) = \exp(h^L_k)/{\sum_{j=1}^m\exp(h^L_j)}$, $k=1,\cdots,m$.
The neural network parameters $\bm\theta$ consists of $\bm{W}^l$ and $\bm{b}^l$, $l=1,\cdots,L$.
In this supervised setting, the code length in \cref{eq:bayes} means the predictive loss of the Bayesian mixture model $p(\bm{x})= p(\bm{z}) \int p(y\mid\bm{z},\bm\theta)p(\bm\theta)\dtheta$ \wrt the observed pairs $(\bm{z}_i,y_i)$. The smaller the code length, the more accurate the prediction.

All such neural networks
$\mathrm{NN}_{\bm\theta}$ when $\bm\theta$ varies in a parameter space
are referred to as the \emph{neuromanifold}:
$\calM_{\bm\theta}=\left\{ \mathrm{NN}_{\bm\theta}:\;{\bm\theta\in\Theta} \right\}$.
Similarly, the parameter space of the distribution family $p(\bm{z} \mid \bm\psi)$
is denoted as $\calM_{\bm\psi}$.
In machine learning, we are often interested in the FIM \wrt $\bm\theta$
as it reveals the geometry of the parameter space.
However, the FIM can also be computed relatively \wrt a subset of
$\bm\theta$ in a sub-system~\citep{snRFI}.

By the definition in \cref{eq:fim}, the FIM on the product manifold
$\calM_{\bm\psi}\times\calM_{\bm\theta}$ is in a block-diagonal form
\begin{equation}
    \fim(\bm\psi,\bm\theta)
    =
\begin{bmatrix}
    \fim(\bm\psi) & \bm0\\
    \bm0 & \fim(\bm\theta)\\
\end{bmatrix}.
\end{equation}
\rev{The off-diagonal blocks are zero-matrices (denoted as $\bm0$) because the generative and predictive models do not share parameters. Indeed, we have}
\begin{align*}
&\expect_p\left(
    \frac{\partial\log p(\bm{x} \mid \bm\psi, \bm\theta)}{\partial\bm\psi}
    \frac{\partial\log p(\bm{x} \mid \bm\psi, \bm\theta)}{\partial\bm\theta^\top}
    \right)
=
\expect_p\left( \frac{\partial\log p(\bm{z} \mid \bm\psi)}{\partial\bm\psi}
\frac{\partial\log p(y \mid \bm{z},\,\bm\theta)}{\partial\bm\theta^\top} \right)\\
&=
\expect_{p(\bm{z} \mid \bm\psi)}
\left( \frac{\partial\log p(\bm{z} \mid \bm\psi)}{\partial\bm\psi}
    \expect_{p(y \mid \bm{z},\,\bm\theta)}
    \left(
\frac{\partial\log p(y \mid \bm{z},\,\bm\theta)}{\partial\bm\theta}
\right)^\top
\right)
=
\bm{0},
\end{align*}
where $\expect_{p(y \mid \bm{z},\,\bm\theta)} \left(
\frac{\partial\log p(y \mid \bm{z},\,\bm\theta)}{\partial\bm\theta}
\right)$ is the expectation of the score function and is always zero.
The metric $\fim(\bm\psi, \bm\theta)$ is a \emph{product metric},
meaning that the geometry of $\calM_{\bm\theta}$ defined by $\fim(\bm\theta)$ can be studied
separately to the geometry of $\calM_{\bm\psi}$.

As we are interested in the predictive model corresponding to the diagonal block $\fim(\bm\theta)$, we further have
(see \eg~\citep{pbRNG}\citep{varfim} for derivations)
\begin{align}\label{eq:j}
\fim(\bm\theta)
&=
\expect_{p(\bm{z})}
\left[
\left( \frac{\partial\bm{h}^L(\bm{z})}
{\partial\bm\theta} \right)^\top
\;\bm{C}(\bm{z})\;
\frac{\partial\bm{h}^L(\bm{z})}{\partial\bm\theta}
\right],
\end{align}
where the expectation is taken \wrt $p(\bm{z})\defeq{}p(\bm{z}\mid\bm\psi)$, an underlying true
distribution in the input space depending on the parameter $\bm\psi$.
$\frac{\partial\bm{h}^L(\bm{z})}{\partial\bm\theta}$ is the $m\times{D}$
parameter-output Jacobian matrix, based on a given input $\bm{z}$,
$\bm{C}(\bm{z})\defeq\diag{\bm{o}(\bm{z})}-\bm{o}(\bm{z})\bm{o}(\bm{z})^\top\succeq0$,
$\diag{\cdot}$ means the diagonal matrix with the given diagonal entries,
and $\bm{o}(\bm{z}) \defeq \softmax(\bm{h}^L(\bm{z}))$
is the predicted class probabilities of $\bm{z}$.
By the definition of $\softmax$, each dimension of
$\bm{o}(\bm{z})$ represents a positive probability, although $\bm{o}(\bm{z})$ can be arbitrarily close to a one-hot vector.
As a result, the kernel of the psd matrix $\bm{C}(\bm{z})$ is given by $\{\lambda \bm{1}\,:\,\lambda\in\Re \}$,
where $\bm{1}$ is the vector of all $1$'s.

In \cref{eq:j}, $\fim(\bm\theta)$ is the single-observation FIM.
It is obvious that the FIM \wrt the joint distribution $p(\bm{X}\,\vert\,\bm\theta)$
of multiple observations is $N\fim(\bm\theta)$ (Fisher information is additive),
so that $\fim(\bm\theta)$ does not scale with $N$. In theory, computing $\fim(\bm\theta)$ requires assuming $p(\bm{z})$, which depends on the parameter $\bm\psi$. This makes sense as $(\bm\psi_1,\bm\theta)$ and $(\bm\psi_2, \bm\theta)$ with $\bm\psi_1\neq\bm\psi_2$ are
different points on the product manifold $\calM_{\bm\psi}\times\calM_{\bm\theta}$
and thus their $\fim(\bm\theta)$ should be different.
In practice, one only gets access to
a set of $N$ \iid samples drawn from an unknown $p(\bm{z}\mid\bm\psi)$.
In this case, it is reasonable to take $p(\bm{z})$ in \cref{eq:j}
to be the empirical distribution $\hat{p}(\bm{z})$ so that
$p(\bm{z}) = \hat{p}(\bm{z})\defeq\frac{1}{N}\sum_{i=1}^N\delta(\bm{z}-\bm{z}_i)$, then
\begin{align}\label{eq:j2}
\fim(\bm\theta)
=
\hat{\fim}(\bm\theta)
&\defeq
\frac{1}{N} \sum_{i=1}^N
\left[
\left( \frac{\partial\bm{h}^L(\bm{z}_i)}{\partial\bm\theta} \right)^\top
\;\bm{C}(\bm{z}_i)\;
\frac{\partial\bm{h}^L(\bm{z}_i)}{\partial\bm\theta}
\right].
\end{align}
\rev{The FIM computed in this way does not rely on the assumption of a parametric generative model $p(\bm{z}\mid\bm\psi)$ and the choice of a $\bm\psi$.}
$\hat{\fim}(\bm\theta)$
can be directly computed from the observed $\bm{z}_i$'s and does not depend on the observed $y_i$'s.
Although denoted differently than $\fim(\bm\theta)$ in the current paper, this $\hat{\fim}(\bm\theta)$
is a standard version of the definition of the FIM for neural networks~\cite{martens,limit,snRFI,varfim}.

By considering the neural network weights and biases as random variables
satisfying a prescribed prior distribution~\cite{kaUSO,pwTSO}, this
$\fim(\bm\theta)$ can be regarded as a random matrix~\cite{msFPA} depending on
the structure of the DNN and the prior.
The empirical density of $\fim(\bm\theta)$ is the empirical distribution of its eigenvalues $\{\lambda_i\}_{i=1}^D$,
that is, $\rho_D(\lambda) = \frac{1}{D}\sum_{i=1}^D \delta(\lambda_i)$.
If at the limit $D\to\infty$, the empirical density converges to a probability density function (pdf), then
\begin{equation}
\rho_{\fim}(\lambda) \defeq \lim_{D\to\infty} \rho_D(\lambda)
\end{equation}
is called the \emph{spectral density} of the Fisher information matrix.

For DNN, we assume that
\begin{description}
    \item[(\text{\bf{A1}})]
        At the MLE $\hat{\bm\theta}$, the prediction $\softmax(\bm{h}^L(\bm{z}_i))$
        perfectly recovers (tending to be one-hot vectors) the training target $y_i$,
        for all the training samples $(\bm{z}_i, y_i)$.
\end{description}
In this case, the negative Hessian of the average log-likelihood
\begin{equation*}
\ofim(\bm\theta)
\defeq
-\frac{1}{N}\frac{\partial^2 \log p(\bm{X}\,\vert\,\bm\theta)}
                   {\partial \bm\theta\partial\bm\theta^\top}
=
-
\frac{1}{N}\sum_{i=1}^N
\frac{\partial^2 \log p(y_i\,\vert\,\bm{z}_i,\bm\theta)}
{\partial\bm\theta\partial\bm\theta^\top}
\end{equation*}
is called the observed FIM (sample-based FIM), which is also known as the ``empirical Fisher'' in machine learning literature~\cite{martens,limit}.
In our notations explained in \cref{tbl:fim}, the FIM $\fim$ depends on the
true distribution $p(\bm{z})$
and does not depend on the observed samples.
In the expression of the FIM in \cref{eq:j},
if $p(\bm{z})=\hat{p}(\bm{z})$,
then $\fim$ become $\hat{\fim}$,
which depends on the observed
input $\bm{z}_i$'s.
The observed FIM
$\ofim$ depends on
both the observed input $\bm{z}_i$'s and the observed target $y_i$'s.
If $p(\bm{z})=\hat{p}(\bm{z})$,
the observed FIM coincides with the FIM at the MLE $\hat{\bm\theta}$ and
$\ofim(\hat{\bm\theta})=\hat{\fim}(\hat{\bm\theta})$.
For general statistical models, there is a residual term in between these two matrices which scales
with the training error~(see \eg Eq. 6.19 in section 6 of \cite{aoDOL}, or
\cref{eq:hess} in the appendix).
How these different metric tensors are called is just a matter of terminology. One should distinguish them by examining whether/how they depend (partially) on the observed information.

\begin{table}[ht]
\centering
\caption{The FIM and the observed FIM. The last three columns explain whether the tensor depends on the observed $\bm{z}_i$'s, whether it depends on the observed $y_i$'s, and whether they can be computed in practice based on empirical observations.}\label{tbl:fim}
\begin{tabularx}{.95\textwidth}{rlccc}
\toprule
Notation & Name & Depend on $\bm{z}_i$ & Depend on $y_i$ & Computable\\
\midrule
$\fim(\bm\theta)$ & FIM (\wrt true $p(\bm{z})$) & No & No & No \\
$\hat{\fim}(\bm\theta)$ & FIM (\wrt empirical $\hat{p}(\bm{z})$)& Yes & No & Yes\\
$\ofim(\bm\theta)$ & observed FIM & Yes & Yes & Yes\\
\bottomrule
\end{tabularx}
\end{table}

\section{Local Dimensionality}\label{sec:localdim}

This section quantitatively measures the singularity of the neuromanifold.
Our main definitions and results do \emph{not} depend on the settings introduced in the previous section
and can be generalized to similar models including stochastic neural networks~\citep{calin2020deep}.
For example, if the output units or the network structure is changed, the expression of the FIM and related results can be adapted straightforwardly.
Our derivations depend on that (1) DNNs have a large amount of singularity
corresponding zero eigenvalues of the FIM;
and (2) the spectrum of the (observed) FIM has many eigenvalues close to zero~\cite{PathologicalSpectrumFIM-2021}.
That being said, our results also apply to singular models~\cite{wAGA} with similar properties.

\begin{definition}[Local dimensionality]
    The local dimensionality $d(\bm\theta)\defeq\rank{\fim(\bm\theta)}$ of
    the neuromanifold $\calM$
    at $\bm\theta\in\calM$ refers to the rank of the FIM $\fim(\bm\theta)$.
    If $p(\bm{z})=\hat{p}(\bm{z})$, then
    $d(\bm\theta)=\hat{d}(\bm\theta)\defeq\rank{\hat{\fim}(\bm\theta)}$.
\end{definition}

The local dimensionality $d(\bm\theta)$ is the number of degrees of freedom at
$\bm\theta\in\calM$ which can change the probabilistic model
$p(y\,\vert\,\bm{z},\bm\theta)$ in terms of information theory.
One can find a reparameterized DNN with $d(\bm\theta)$ parameters,
which is locally equivalent to the original DNN with $D$ parameters.
Recall the dimensionality of the tangent bundle is two times the dimensionality of the
manifold.
\begin{remark}
The dimensionality of the screen distribution $\scr$ at $\bm{\theta}$ is $2\, d(\bm\theta)$.
\end{remark}

By definition, the FIM as the singular semi-Riemannian metric of $\calM$ must be psd.
Therefore it only has positive and zero eigenvalues, and the number of positive
eigenvalues $d(\bm\theta)$ is not constant as ${\bm\theta}$ varies in general.
\begin{remark}\label{thm:signature}
The local metric signature (number of positive, negative, zero eigenvalues of the FIM)
of the neuromanifold $\calM$ is $\left(d(\bm\theta),0,D-d(\bm\theta)\right)$,
where $d(\bm\theta)$ is the local dimensionality.
\end{remark}

The local dimensionality $d(\bm\theta)$ depends on the specific choice of $p(\bm{z})$.
If $p(\bm{z})=\hat{p}(\bm{z})$, then
$d(\bm\theta)=\hat{d}(\bm\theta)=\rank{\hat{\fim}(\bm\theta)}$.
On the other hand, one can use the rank of the negative Hessian $\ofim(\bm\theta)$ (\ie, observed rank)
to get an approximation of the local dimensionality $d(\bm\theta)\approx
\rank{\ofim(\bm\theta)}$. In the MLE $\hat{\bm\theta}$, this approximation becomes accurate. We simply denote $d$ and $\hat{d}$, instead of
$d(\bm\theta)$ and $\hat{d}(\bm\theta)$, if $\bm\theta$ is clear from the context.

We first show that the lightlike dimensions of $\calM$ do not affect the neural
network model in \cref{eq:nn}.
\begin{lemma}\label{thm:kernel}
    If $(\bm\theta, \sum_{j}\alpha_j\partial\theta_j)\in\rad$,
    \ie $\langle\sum_{j}\alpha_j\partial\theta_j,
                \sum_{j}\alpha_j\partial\theta_j\rangle_{\fim(\bm\theta)} = 0$,
    then almost surely we have
    $\frac{\partial\bm{h}^L(\bm{z})}{\partial\bm\theta} \bm\alpha = \lambda(\bm{z}) \bm{1}$,
    where $\lambda(\bm{z})\in\Re$.
\end{lemma}
By \cref{thm:kernel},
the Jacobian $\frac{\partial\bm{h}^L(\bm{z})}{\partial\bm\theta}$ is the local
linear approximation of the map $\bm\theta\to\bm{h}^L$.  The dynamic $\bm\alpha$
(coordinates of a tangent vector) on $\calM$ causes a uniform increment on the
output $\bm{h}^L$, which, after the $\softmax$ function, does not change the
neural network map $\bm{z}\to{y}$.

\rev{Then, we can upper-bound the local dimensionality using the rank of the parameter-output Jacobian $\partial \bm{h}^L(\bm{z})/\partial\bm\theta$.}
\begin{proposition}\label{thm:dim}
\rev{$\forall\bm\theta\in\calM$,
  $\hat{d}(\bm\theta) \le
  \sum_{i=1}^N
\min\left\{
\rank{\frac{\partial \bm{h}^L(\bm{z}_i)}{\partial\bm\theta}},
m-1 \right\}.$}
\end{proposition}

\begin{remark}
  While the total number $D$ of free parameters is unbounded in DNNs, the local
  dimensionality estimated by $\hat{d}(\bm\theta)$ grows at most linearly \wrt the sample size $N$,
  given fixed $m$ (size of the last layer).
  If both $N$ and $m$ are fixed, then $\hat{d}(\bm\theta)$ is bounded even when
  the network width $M\to\infty$ and/or depth $L\to\infty$.
\end{remark}

\rev{The above bound is based on the inequality $\rank{\sum_{i}\bm{A}_i} \le \sum_i\rank{\bm{A}_i}$ for any matrices $\bm{A}_i$, which could lead to loose bounds.
Alternatively, an upper bound can be established directly based on the definition of the matrix rank.}
\begin{proposition}\label{thm:dim2}
\rev{For all $\bm\theta\in\calM$, we have
$d(\bm\theta)
\le
\dim\mathrm{span}
\left(
\bigcup_{\bm{z}\in\mathrm{supp}(p)}
\mathrm{Row}\left( \frac{\partial\bm{h}^L(\bm{z})}{\partial\bm\theta} \right)
\right),
$
where $\mathrm{supp}(p)$ is the support of $p(\bm{z})$, and
$\mathrm{Row}(\bm{A})$ denotes the row vectors of the matrix $\bm{A}$.
Similarly, $\hat{d}(\bm\theta)$ has an upper bound obtained by replacing $\mathrm{supp}(p)$ with $\mathrm{supp}(\hat{p})$ on the RHS, \ie the union is over the observed $\bm{z}_i$'s.}
\end{proposition}
\rev{In summary, the less the rank of $\frac{\partial\bm{h}^L(\bm{z})}{\partial\bm\theta}$, the more potential singularities in the neuromanifold.
Note the Jacobian $\frac{\partial\bm{h}^L(\bm{z})}{\partial\bm\theta}$ can be further written as
\begin{equation}
\frac{\partial\bm{h}^L(\bm{z})}{\partial\bm\theta}
=
\begin{pmatrix}
\frac{\partial\bm{h}^L(\bm{z})}{\partial\bm{w}^1},
\frac{\partial\bm{h}^L(\bm{z})}{\partial\bm{w}^2},
\cdots,
\frac{\partial\bm{h}^L(\bm{z})}{\partial\bm{w}^L}
\end{pmatrix},
\end{equation}
where $\bm{w}^l$ contains all parameters in the $l$'th layer ($l=1,\cdots,L$) and is obtained by stacking the columns of $\bm{W}^l$ and $\bm{b}^l$ into a long vector.
We can bound $\frac{\partial\bm{h}^L(\bm{z})}{\partial\bm{w}^l}$ individually as below.}
\begin{proposition}\label{thm:dim3}
\rev{It holds that
\begin{align}
\rank{\frac{\partial\bm{h}^L(\bm{z})}{\partial\bm{w}^l}}
\le
\rank{ \bm{W}^L \bm\Phi^{L-1} \bm{W}^{L-1} \cdots \bm{W}^{l+1} \bm\Phi^l}
\le
\min_{s=l}^{L-1} \rank{\bm\Phi^s},
\end{align}
where $\bm{\Phi}^l=\diag{\phi'(\bm{h}^l_1), \cdots, \phi'(\bm{h}^l_M)}$ is the Jacobian of the $l$'th activation layer.}
\end{proposition}
\rev{
Observe that the upper bounds in \cref{thm:dim3} are monotonically decreasing with respect to $L$. For example, we have
\begin{equation}
\rank{ \bm{W}^L \bm\Phi^{L-1} \cdots \bm{W}^{l+1} \bm\Phi^l}
\le
\rank{ \bm{W}^L \bm\Phi^{L-1} \cdots \bm{W}^{l+2} \bm\Phi^{l+1}}.
\end{equation}
Based on these upper bounds,
$\frac{\partial\bm{h}^L(\bm{z})}{\partial\bm{w}^l}$ is potentially more singular for layers that are close to the input.
\begin{remark}
If $\phi$ is ReLU, then the diagonal entries of $\bm\Phi^s$ form a binary vector, and $\rank{\bm\Phi^s}$ is the number of activated neurons in the $s$'th layer. In this case,
the upper bound $\min_{s=l}^{L-1} \rank{\bm\Phi^s}$
means the smallest number of activated neurons across all layers.
\end{remark}
}

To understand $d(\bm\theta)$,
one can parameterize the DNN, locally, with only $d(\bm\theta)$ free parameters
while maintaining the same predictive model. The log-likelihood is a function of
these $d(\bm\theta)$ parameters, and therefore its Hessian has at most rank
$d(\bm\theta)$.
In theory, one can only reparameterize $\calM$ so that at one single point $\hat{\bm\theta}$,
the screen and radical distributions are separated based on the coordinate chart.
Such a chart may neither exist locally (in a neighborhood around $\hat{\bm\theta}$) nor globally.

The local dimensionality is not constant and may vary with $\bm\theta$.
The global topology of the neuromanifold is therefore like a \emph{stratifold}~\citep{akOTC,EsserNielsen}.
As $\bm\theta$ has a large dimensionality in DNNs, singularities are more likely to occur in $\calM$.
Compared to the notion of \emph{intrinsic dimensionality}~\citep{lfMTI},
our $d(\bm\theta)$ is well-defined mathematically rather than based on empirical
evaluations. One can regard our local dimensionality as an upper bound of the
intrinsic dimensionality, because a very small singular value of $\fim$ still
counts towards the local dimensionality. Notice that random matrices have full
rank with probability 1~\citep{fzTRO}.

We can regard small singular values
(below a prescribed threshold $\varepsilon>0$) as $\varepsilon$-singular dimensions, and use $\varepsilon$-rank defined below to estimate the local dimensionality.
\begin{definition}
    The $\varepsilon$-rank of the FIM $\fim(\bm\theta)$ is the number of eigenvalues of $\fim(\bm\theta)$ which is not less than some given $\varepsilon>0$.
\end{definition}
\noindent{}By definition, the $\varepsilon$-rank is a lower bound of the rank of the FIM,
which depends on the $\bm\theta$-parameterization --- different parameterizations of the DNN may yield different
$\varepsilon$-ranks of the corresponding FIM.
If $\varepsilon\to0$, the $\varepsilon$-rank of $\fim(\bm\theta)$ becomes the true rank of $\fim(\bm\theta)$ given by $d(\bm\theta)$.
The spectral density $\rho_{\fim}$ (probability distribution of the eigenvalues
of $\fim(\bm\theta)$) affects the $\varepsilon$-rank of $\fim(\bm\theta)$
and the expected local dimensionality of $\calM$.
On the support of $\rho_{\fim}$, the higher the probability of the region $[0,\varepsilon)$,
the more likely $\calM$ is singular.
By the Cram\'er-Rao lower bound, the variance of an unbiased 1D estimator $\hat\theta$ must satisfy
\begin{equation*}
\mathrm{var}(\hat{\theta})\ge\fim(\theta)^{-1}\ge\frac{1}{\varepsilon}.
\end{equation*}
Therefore the $\varepsilon$-singular dimensions lead to
a large variance of the estimator $\hat{\theta}$: a single observation
$\bm{x}_i$ carries little or no information regarding $\theta$,
and it requires a large number of observations to achieve the same precision.
The notion of thresholding eigenvalues close to zero may depend on
the parameterization but the intrinsic ranks given by the local dimensionality are invariant.

In a DNN, there are several typical \emph{sources of singularities}:
\begin{itemize}
\item First, if
a neuron is saturated and gives constant output regardless of the input sample
$\bm{z}_i$, then all dynamics of its input and output connections are in $\rad$.

\item Second, two neurons in the same layer can have linearly dependent output,
\eg when they share the same weight vector and bias.
They can be merged into one single neuron, as there exists redundancy in the
original parameterization.

\item Third, if the activation function $\phi(\cdot)$
is homogeneous, \eg ReLU, then any neuron in the DNN induces a reparametrization
by multiplying the input links by $\alpha$ and output links by $1/\alpha^k$ ($k$
is the degree of homogeneity). This reparametrization corresponds to a null
curve in the neuromanifold parameterized by $\alpha$.

\item Fourth, certain structures
such as recurrent neural networks (RNNs) suffer from vanishing gradient~\cite{goodfellow2016deep}. As the
FIM is the variance of the gradient of the log-likelihood (known as variance of the score in statistics), its scale goes to
zero along the dimensions associated with such structures.
\end{itemize}

It is meaningful to formally define the notion of ``lightlike neuromanifold''.
Using geometric tools, related studies can be invariant \wrt neural network reparametrization.
Moreover, the connection between neuromanifold and singular semi-Riemannian geometry,
which is used in general relativity, is not yet widely adopted in machine learning.
For example, the textbook~\cite{wAGA} in singular statistics
mainly used tools from algebraic geometry which is a different field.

Notice that the Fisher-Rao distance along a null curve is undefined because
there the FIM is degenerate and there is no arc-length reparameterization
along null curves~\citep{kay1998}.

\section{General Formulation of Our Razor}\label{sec:general}
In this section, we derive a new formula of MDL for DNNs, aiming to explain
\emph{how does the high dimensional DNN structure can have a short code length}
of the given data?
Notice that, this work focuses on the concept of model complexity but not the
generalization bounds. We aim to show that the DNN model is intrinsically simple
because it can be described shortly.
The theoretical connection between generalization power and MDL is studied
in PAC-Bayesian theory and PAC-MDL (see \cite{grMDL,MDLbarroncover,MDLzhang,pmlr-v98-grunwald19a,PACMDLbounds} and references
therein). This is beyond the scope of this paper.

We derive a simple asymptotic formula for the case of large sample size
and large network size.
Therefore crude approximations are taken and the low-order terms are ignored,
which are common practices in deriving information criteria~\citep{aANL,sETD}.

In the following, we will abuse $p(\bm{x}\mid\bm\theta)$ to denote the
DNN model $p( y \mid \bm{z},\,\bm\theta)$ for shorter equations and
to be consistent with the introduction.
Assume
\begin{description}
    \item[(\text{\bf{A2}})]
        The absolute values of the third-order derivatives of
        $\log{p}(\bm{x}\,\vert\,\bm\theta)$ \wrt $\bm\theta$ are bounded by some constant.
   \item[(\text{\bf{A3}})]
        $\forall{i}$, $\vert \theta_i  - \hat{\theta}_i\vert = O(1/\sqrt{M})$,
        where $O(\cdot)$ is Bachmann–Landau's big-O notation.
\end{description}
Recall that $M$ is the width of the neural network. We consider that
the NN weights have a order of $O(1/\sqrt{M})$. For example, if the
input to a neuron follows the standard Gaussian distribution, then its weights
with order $O(1/\sqrt{M})$ guarantee the output is $O(1)$. In practice, this
constraint can be enforced by clipping the weight vector to a prescribed range.
\rev{This scaling is commonly used by random initialization techniques~\cite{glorot2010understanding,he2015delving} for training DNNs.}

We rewrite the code length in \cref{eq:bayes} based on the Taylor expansion of
$\log{p}(\bm{X}\mid\bm\theta)$
at $\bm\theta=\hat{\bm\theta}$ up to the second order:
\begin{align}\label{eq:px}
-\log p(\bm{X}) = &
-\log \int_{\calM} p(\bm\theta)
\exp\bigg( \log{p}(\bm{X}\,\vert\,\hat{\bm\theta})
-\frac{N}{2} (\bm\theta-\hat{\bm\theta})^\top
\ofim(\hat{\bm\theta}) (\bm\theta-\hat{\bm\theta})\nonumber\\
&\quad\quad+O\left(N\Vert\bm\theta-\hat{\bm\theta}\Vert^3\right)
\bigg) \dtheta.
\end{align}
Notice that the first order term vanishes because $\hat{\bm\theta}$ is a local
optimum of $\log p(\bm{X}\,\vert\,\bm\theta)$, and in the second order term,
$-N\ofim(\hat{\bm\theta})$ is the Hessian matrix of
the likelihood function $\log{p}(\bm{X}\,\vert\,\bm\theta)$ evaluated at $\hat{\bm\theta}$.
At the MLE, $\ofim(\hat{\bm\theta})\succeq0$, while in general the Hessian of the loss of a DNN
evaluated at $\bm\theta\neq\hat{\bm\theta}$ can have a negative spectrum~\citep{alNEO,seEAO}.

Through a change of variable $\bm\phi\defeq\sqrt{N}(\bm\theta-\hat{\bm\theta})$,
the density of $\bm\phi$ is
$p(\bm\phi)=\frac{1}{\sqrt{N}}p(\frac{\bm\phi}{\sqrt{N}}+\hat{\bm\theta})$
so that $\int_{\calM}p(\bm\phi)d\bm\phi=1$.
In the integration in \cref{eq:px}, the term
$-\frac{N}{2} (\bm\theta-\hat{\bm\theta})^\top \ofim(\hat{\bm\theta}) (\bm\theta-\hat{\bm\theta})$
has an order of $O(\Vert\bm\phi\Vert^2)$.
The cubic remainder term has an order of $O(\frac{1}{\sqrt{N}}\Vert\bm\phi\Vert^3)$.
If $N$ is sufficiently large, this remainder can be ignored. Therefore we can write
\begin{align}\label{eq:simplepx}
-\log{p}(\bm{X})
    &\approx
- \log{p}(\bm{X}\,\vert\,\hat{\bm\theta}) - \log \expect_p \exp\bigg( -\frac{N}{2} (\bm\theta-\hat{\bm\theta})^\top
\ofim(\hat{\bm\theta}) (\bm\theta-\hat{\bm\theta}) \bigg).
\end{align}
On the RHS, the first term measures the error of the model \wrt the observed
data $\bm{X}$. The second term measures the model complexity.
We have the following bound.
\begin{proposition}\label{thm:general}
We have $\forall\bm\theta\in\calM$,
\begin{align*}
0
&\le{}
 - \log \expect_p \exp\bigg(
     -\frac{N}{2} (\bm\theta-\hat{\bm\theta})^\top
     \ofim(\hat{\bm\theta}) (\bm\theta-\hat{\bm\theta}) \bigg)
     \nonumber\\
&\le{}
\frac{N}{2} \trace\left(
\ofim(\hat{\bm\theta})
\left( (\mean{\bm\theta}-\hat{\bm\theta})
(\mean{\bm\theta}-\hat{\bm\theta})^\top + \cov{\bm\theta} \right)
\right),
\end{align*}
where $\mean{\bm\theta}$ and $\cov{\bm\theta}$ denote the mean and covariance
matrix of the prior $p(\bm\theta)$, respectively.
\end{proposition}
Therefore the complexity is always non-negative and its scale is bounded by the
prior $p(\bm\theta)$. The model has low complexity when $\hat{\bm\theta}$ is
close to the mean of $p(\bm\theta)$ and/or when the variance of $p(\bm\theta)$ is small.

Consider the prior
$p(\bm\theta)= \kappa(\bm\theta) / \int_{\calM} \kappa(\bm\theta) d\bm\theta$,
where $\kappa(\bm\theta)>0$ is a positive measure on $\calM$
so that $0<\int_{\calM}\kappa(\bm\theta)\dtheta < \infty$. Based on the above
approximation of $-\log p(\bm{X})$, we arrive at a general formula
\begin{align}\label{eq:general}
\calO
\defeq
    & - \log{p}(\bm{X}\,\vert\,\hat{\bm\theta})
      + \log\int_{\calM}\kappa(\bm\theta)\dtheta \nonumber\\
    & - \log\int_{\calM}
        \kappa(\bm\theta) \exp\left( -\frac{N}{2} (\bm\theta-\hat{\bm\theta})^\top
        \ofim(\hat{\bm\theta}) (\bm\theta-\hat{\bm\theta})
    \right) \dtheta,
\end{align}
where ``$\calO$'' stands for Occam's razor. Compared with previous formulations of MDL~\cite{rMBS,rFIA,bMDL}, \cref{eq:general} relies on a quadratic approximation of the log-likelihood function and can be instantiated based on different assumptions of $\kappa(\bm\theta)$.
\rev{The non-normalized $\kappa(\bm\theta)$ in Bayesian coding serves a similar role to the luckiness function in NML coding~\cite{pTMDL}, as they both incorporate prior knowledge to favor certain parameters in the parameter space $\calM$.}

Informally, the term $\int_{\calM}\kappa(\bm\theta)\dtheta$ gives the total
capacity of models in $\calM$ specified by the \emph{improper prior} $\kappa(\bm\theta)$,
up to constant scaling.
For example, if $\kappa(\bm\theta)$ is uniform on a subregion in $\calM$,
then $\int_{\calM}\kappa(\bm\theta)\dtheta$ corresponds to the size of this region
\wrt the base measure $\dtheta$.
The term $\int_{\calM} \kappa(\bm\theta) \exp\left(
        -\frac{N}{2} (\bm\theta-\hat{\bm\theta})^\top
        \ofim(\hat{\bm\theta}) (\bm\theta-\hat{\bm\theta}) \right) \dtheta$
gives the model capacity specified by the posterior
$p(\bm\theta\,\vert\,\bm{X})
\propto{}
p(\bm\theta)p(\bm{X}\,\vert\,\bm\theta)
\propto
\kappa(\bm\theta) \exp\left(
        -\frac{N}{2} (\bm\theta-\hat{\bm\theta})^\top
        \ofim(\hat{\bm\theta}) (\bm\theta-\hat{\bm\theta}) \right)$.
It shrinks to zero when the number $N$ of observations increases.
The last two terms in \cref{eq:general} is the log-ratio between
the model capacity \wrt the prior and the capacity \wrt the posterior.
A large log-ratio means
there are many distributions on $\calM$ which have a relatively large value of $\kappa(\bm\theta)$
but a small value of
$\kappa(\bm\theta) \exp\left( -\frac{N}{2} (\bm\theta-\hat{\bm\theta})^\top
\ofim(\hat{\bm\theta}) (\bm\theta-\hat{\bm\theta})\right)$.
The associated model is considered to have a high complexity,
meaning that only a small ``percentage'' of the models are
helpful to describe the given data.

DNNs have a large amount of \emph{symmetry}: the parameter space consists of many
pieces that look exactly the same. This can be caused \eg by permutating the
neurons in the same layer. This is a different \emph{non-local} property than
singularity that is a local differential property. Our $\calO$ is not
affected by the model size caused by symmetry, because these symmetric models
are both counted in the prior and the posterior, and the log-ratio in
\cref{eq:general} cancels out symmetric models.  Formally, $\calM$ has
$\zeta$ symmetric pieces denoted by $\calM_1, \cdots,
\calM_{\zeta}$.  Note any MLE on $\calM_i$ is mirrored on those
$\zeta$ pieces.  Then both integrations on the RHS of \cref{eq:general} are
multiplied by a factor of $\zeta$.  Therefore $\calO$ is invariant to
symmetry.

\section{Connection with \texorpdfstring{$f$}{f}-mean}\label{sec:fmean}

\rev{In this section, we show that the model complexity terms in \cref{eq:simplepx,eq:general} can be studied based on the notion of $f$-mean, defined below.}

\begin{definition}[$f$-mean]
    Given a set $\mathfrak{T}=\{t_i\}_{i=1}^n\subset{}\Re$
    and a continuous and strictly monotonous function $f\,:\,\Re\to\Re$,
    the $f$-mean of $\mathfrak{T}$ is
\begin{equation*}
    M_f(\mathfrak{T})
    \defeq
    f^{-1}\left(
        \frac{1}{n}\sum_{i=1}^n f(t_i)
        \right).
\end{equation*}
\end{definition}
The $f$-mean, also known as the quasi-arithmetic mean was
studied in~\citep{Nagumo-1930,kolmogorov1930notion}:
Thus they are also called Kolmogorov-Nagumo means~\cite{komori2021unified}.
By definition, the image of $M_f(\mathfrak{T})$ under $f$ is the arithmetic mean of the
image of $\mathfrak{T}$ under the same mapping.
Therefore, $M_f(\mathfrak{T})$ is in between the smallest and largest elements of $\mathfrak{T}$.
If $f(x)=x$, then $M_f$ becomes the arithmetic mean, which we denote as $\overline{\mathfrak{T}}$.
We have the following bound.
\begin{lemma}\label{thm:b}
    Given a real matrix $\bm{T}=(t_{ij})_{n\times{m}}$,
    we use $\bm{t}_i$ to denote the $i$'th row of $\bm{T}$,
    and $\bm{t}_{:,j}$ to denote the $j$'th column of $\bm{T}$.
    If $f(t)=\exp(-t)$, then
    \begin{equation*}
        M_f(\bm{T})
        \le
        \overline{\left\{M_f(\bm{t}_{:,1}), \cdots, M_f(\bm{t}_{:,m})\right\}}
        \le
        M_f( \{\overline{\bm{t}_1}, \cdots, \overline{\bm{t}_n} \} )
        \le
        \overline{\bm{T}},
    \end{equation*}
    where $M_f(\bm{T})$ is the $f$-mean of all $n\times{m}$ elements of $\bm{T}$,
    and $\overline{\bm{T}}$ is their arithmetic mean.
\end{lemma}
\rev{Particular attention should be given to the second ``$\le$''.}
If the arithmetic mean of each row is first evaluated,
and then their $f$-mean is evaluated, we get an upper bound of the arithmetic
mean of the $f$-mean of the columns.
In simple terms, for $f(t)=\exp(-t)$, \emph{the $f$-mean of arithmetic mean is lower bounded by the arithmetic mean of the $f$-mean}.
The proof is straightforward from Jensen's inequality, and by noting that
$-\log\sum_{i}\exp(-t_i)$ is a concave function of $\bm{t}$.
The last ``$\le$'' leads to a proof of the upper bound in \cref{thm:general}.
\rev{\Cref{thm:b} pertains to the mean of a matrix. It leads to bounds of the mean value of a bi-variable function \wrt a probability mass function or a probability density function, or a mix of both. This is straightforward and omitted.}

\begin{remark}
\rev{All instances of ``$\le$'' in \cref{thm:b} are derived from Jensen's inequality. Consequently, the gaps of these bounds
shrink as the variance of the matrix elements decreases,
where the specific way of measuring the variance depends on the particular ``$\le$''.
For instance, the gap associated with the second ``$\le$'' becomes smaller as the variance across each row $\bm{t}_i$ reduces.}
\end{remark}

\begin{remark}
The second complexity term on the RHS of \cref{eq:simplepx} is the $f$-mean
of the quadratic term
$\frac{N}{2} (\bm\theta-\hat{\bm\theta})^\top \ofim(\hat{\bm\theta}) (\bm\theta-\hat{\bm\theta})$
\wrt the prior $p(\bm\theta)$,
where $f(t)=\exp(-t)$.
\end{remark}

Based on the spectrum decomposition
$\ofim(\hat{\bm\theta})=\sum_{j=1}^{\rank{\ofim(\hat{\bm\theta})}} \lambda_j^+ \bm{v}_j\bm{v}_j^\top$,
where the positive eigenvalues $\lambda_j^+\defeq\lambda_j^+(\ofim(\hat{\bm\theta}))$
and the eigenvectors $\bm{v}_j\defeq\bm{v}_j(\hat{\bm\theta})$
depend on the MLE $\hat{\bm\theta}$,
we further write this term as
\begin{align*}
\frac{N}{2} (\bm\theta-\hat{\bm\theta})^\top \ofim(\hat{\bm\theta}) (\bm\theta-\hat{\bm\theta})
&=
\sum_{j=1}^{\rank{\ofim(\hat{\bm\theta})}}
\frac{\lambda_j^+}{\trace(\ofim(\hat{\bm\theta}))}
\cdot
\frac{N}{2}\trace(\ofim(\hat{\bm\theta}))
\langle \bm\theta-\hat{\bm\theta}, \bm{v}_j \rangle^2.
\end{align*}
By \cref{thm:b}, we have
\begin{align*}
&     - \log \expect_p \exp\bigg( -\frac{N}{2} (\bm\theta-\hat{\bm\theta})^\top
\ofim(\hat{\bm\theta}) (\bm\theta-\hat{\bm\theta}) \bigg)\nonumber\\
\ge
&
- \sum_{j=1}^{\rank{\ofim(\hat{\bm\theta})}}
\frac{\lambda_j^+}{\trace(\ofim(\hat{\bm\theta}))}
     \log \expect_p \exp\bigg( -\frac{N}{2} \trace(\ofim(\hat{\bm\theta}))
     \langle \bm\theta-\hat{\bm\theta}, \bm{v}_j \rangle^2
\bigg),
\end{align*}
where the $f$-mean and the mean \wrt
$\frac{\lambda_j^+}{\trace(\ofim(\hat{\bm\theta}))}$ is swapped on the RHS.
Denote $\varphi_j=\langle \bm\theta-\hat{\bm\theta}, \bm{v}_j \rangle$, which in matrix form is written as
$\bm\varphi=\bm{V}^\top(\bm\theta-\hat{\bm\theta})$.
$\bm{V}$ has orthonormal columns and the $j$'th column of $\bm{V}$ is $\bm{v}_j$.
The prior of $\bm\varphi$ is given by $p(\bm{V}\bm\varphi+\hat{\bm\theta})$.
\rev{Then
\begin{align*}
-\log \expect_p \exp\bigg( -\frac{N}{2} \trace(\ofim(\hat{\bm\theta}))
\langle \bm\theta-\hat{\bm\theta}, \bm{v}_j \rangle^2
\bigg)
=
-\log \expect_{p(\varphi_j)}\exp\left(
    -\frac{N}{2} \trace(\ofim(\hat{\bm\theta}))
    \varphi_j^2
    \right),
\end{align*}
where the RHS is a variance-like measure of $p(\bm\theta)$ up to scaling, as it is the $f$-mean of
$\frac{N}{2} \trace(\ofim(\hat{\bm\theta})) \varphi_j^2$
evaluated at point $\hat{\bm\theta}$ along the direction $\bm{v}_j$.
In summary, we get a lower bound of the model complexity, which is tighter than the lower bound in \cref{thm:general}, given by
\begin{align}\label{eq:boundfmean1}
&- \log \expect_p \exp\bigg( -\frac{N}{2} (\bm\theta-\hat{\bm\theta})^\top
\ofim(\hat{\bm\theta}) (\bm\theta-\hat{\bm\theta}) \bigg)\nonumber\\
\ge
&
- \sum_{j=1}^{\rank{\ofim(\hat{\bm\theta})}}
\frac{\lambda_j^+}{\trace(\ofim(\hat{\bm\theta}))}
\log \expect_{p(\varphi_j)}\exp\left(
    -\frac{N}{2} \trace(\ofim(\hat{\bm\theta}))
    \varphi_j^2
    \right).
\end{align}
The RHS is determined by the
quantity $\frac{N}{2} \trace\left(\ofim(\hat{\bm\theta})\right)\varphi_j^2$
after evaluating the $f$-mean and some weighted mean, where $\varphi_j$ is an orthogonal transformation
of the local coordinates $\theta_i$ based on the spectrum of $\ofim(\hat{\bm\theta})$.
Recall that the trace of the observed FIM $\ofim(\hat{\bm\theta})$ means the
overall amount of information a random observation contains \wrt the underlying
model. Given the same sample size $N$, a larger
$\trace\left(\ofim(\hat{\bm\theta})\right)$ indicates that
the samples are more informative and the likelihood is
more sensitive to the choice of the parameters on $\calM$.
Consequently, it is reasonable to regard the model as more complex, because small changes of model parameters more easily lead to different representations.}

\rev{The bound in \cref{eq:boundfmean1} is tight when the variance of
$\frac{N}{2} \trace\left(\ofim(\hat{\bm\theta})\right)\varphi_j^2$
\wrt
the discrete distribution
$\frac{\lambda_j^+}{\trace(\ofim(\hat{\bm\theta}))}$ is small.
In the case when $\ofim(\hat{\bm\theta})$ is rank-one, ``$\ge$'' becomes ``$=$''. In practice, the FIM of DNNs exhibits a pathological spectrum~\cite{PathologicalSpectrumFIM-2021}, where most eigenvalues of $\ofim(\hat{\bm\theta})$ are near zero, with a small fraction taking large values.
This means that the $\epsilon$-rank of
$\ofim(\hat{\bm\theta})$ is limited, and
the distribution $\frac{\lambda_j^+}{\trace(\ofim(\hat{\bm\theta}))}$
has lower variance as compared to a uniform spectrum.
This distinctive property of DNNs offers some basis for considering the lower bound in \cref{eq:boundfmean1} as a proxy of the model complexity.}

\rev{As $\hat{\bm\theta}$ is the MLE, we have
$\ofim(\hat{\bm\theta})=\hat{\fim}(\hat{\bm\theta})$. Recall from \cref{eq:j} that the
FIM $\hat{\fim}(\hat{\bm\theta})$ is a numerical average over all observed samples. We
can have alternative lower bounds of the model complexity based on \cref{thm:b}:
\begin{align}\label{eq:lb}
&     - \log \expect_p \exp\bigg( -\frac{N}{2} (\bm\theta-\hat{\bm\theta})^\top
\ofim(\hat{\bm\theta}) (\bm\theta-\hat{\bm\theta}) \bigg)\nonumber\\
\ge
&
-
\frac{1}{N}\sum_{i=1}^N
      \log \expect_p \exp\bigg( -\frac{N}{2} (\bm\theta-\hat{\bm\theta})^\top
\left( \frac{\partial\bm{h}^L(\bm{z}_i)}{\partial\bm\theta} \right)^\top
\bm{C}_i
\frac{\partial\bm{h}^L(\bm{z}_i)}{\partial\bm\theta}
(\bm\theta-\hat{\bm\theta}) \bigg)\nonumber\\
\ge
&
-
\frac{1}{N}\sum_{i=1}^N
\expect_{p(y\mid\bm{z}_i)}
      \log \expect_p \exp\bigg( -\frac{N}{2}
      \left[\frac{\partial\log p(y\mid\bm{z}_i)}{\partial\bm\theta^\top} (\bm\theta-\hat{\bm\theta}) \right]^2
\bigg).
\end{align}
The bounds are obtained by swapping the $f$-mean with the numerical average of the samples, and by swapping $f$-mean with the expectation \wrt $p(y\mid\bm{z}_i)$.
Therefore the model complexity can be bounded by the average scale of the vector
$\frac{\partial\bm{h}^L(\bm{z}_i)}{\partial\bm\theta} (\bm\theta-\hat{\bm\theta})$,
where $\bm\theta\sim p(\bm\theta)$.
Note that $\frac{\partial\bm{h}^L(\bm{z}_i)}{\partial\bm\theta}$
is the parameter-output Jacobian matrix, or a linear approximation
of the neural network mapping $\bm\theta\to\bm{h}^L$.
The complexity lower bounds in \cref{eq:lb} mean how the local parameter change $(\bm\theta-\hat{\bm\theta})$ \wrt the prior $p(\bm\theta)$ affect the output.
If the output is sensitive to
these parameter variations, then the model is considered to have high complexity.
In summary, the $f$-mean offers a powerful tool to analyze our model complexity
and obtain its approximations.}

\section{The Razor based on Gaussian Prior}\label{sec:gauss}

The simplest and most widely-used choice of the prior $p(\bm\theta)$
is the Gaussian prior (see \eg \cite{mackay,kaUSO} among many others).
In \cref{eq:general}, we set
\begin{equation*}
\kappa(\bm\theta)
= \exp\left(-\bm\theta^\top \diag{\frac{1}{\bm\sigma}} \bm\theta\right),
\end{equation*}
where $\diag{\cdot}$ means a diagonal matrix constructed with
given entries, and $\bm\sigma>0$ (elementwisely).
Equivalently, the associated prior is $p_G(\bm\theta) = \calG(\bm\theta\,\vert\,\bm{0}, \diag{\bm\sigma})$,
meaning a Gaussian distribution with mean $\bm{0}$ and covariance matrix $\diag{\bm\sigma}$.
We further assume
\begin{description}
    \item[(\text{\bf{A4}})] $\calM$ has a global coordinate chart and $\calM$ is homeomorphic to $\Re^D$.
    \item[(\text{\bf{A5}})]
    Regardless of $D$, $\hat{\bm\theta}^\top \diag{\frac{1}{\bm\sigma}} \hat{\bm\theta} < \infty$.
\end{description}
\rev{Assumption (A4) enables us to define a Gaussian distribution in a global coordinate system, which typically represents the neural network weights and biases.}
By assumption (A5), the MLE $\hat{\bm\theta}$ has a non-zero probability under the Gaussian prior.

From \cref{eq:general}, we get a closed form expression
of the razor
\begin{align}\label{eq:og}
\calO_G \defeq
   &
    - \log{p}(\bm{X}\,\vert\,\hat{\bm\theta})
    + \frac{\rank{\ofim(\hat{\bm\theta})}}{2} \log N\nonumber\\
   & + \frac{1}{2}
    \sum_{i=1}^{\rank{\ofim(\hat{\bm\theta})}}
    \log \left(
    \lambda_i^+\left(
\ofim(\hat{\bm\theta})\diag{\bm\sigma}\right)
+\frac{1}{N} \right) + O(1),
\end{align}
where
$\lambda_i^+\left( \ofim(\hat{\bm\theta})\diag{\bm\sigma}\right)$ denotes the $i$'th positive eigenvalue of
$\ofim(\hat{\bm\theta})\diag{\bm\sigma}$.
Notice that $\ofim(\hat{\bm\theta})\diag{\bm\sigma}$
and $\diag{\sqrt{\bm\sigma}} \ofim(\hat{\bm\theta})\diag{\sqrt{\bm\sigma}}$
share the same set of non-zero eigenvalues,
and the latter is psd with
$\rank{\ofim(\hat{\bm\theta})}$ positive eigenvalues.

In our razor expressions, all terms that do not scale with the sample size $N$ or the number of parameters $D$ are discarded.
The first two terms on the RHS are similar to BIC~\cite{sETD} up to scaling.
The complexity terms (second and third terms on the RHS of \cref{eq:og}) do not scale with $D$ but are bounded by the rank of
the Hessian, or the observed FIM.
In other words, the radical distribution associated with zero-eigenvalues of $\ofim(\hat{\bm\theta})$ does not affect the model complexity.
This is different from previous formulations of MDL~\citep{rMBS,rFIA,bMDL} and BIC~\cite{sETD}. For example, the 2nd term on the RHS of \cref{eq:mdl} increases linearly with $D$,
\rev{while the 2nd term on the RHS of \cref{eq:og} increases linearly with $\rank{\ofim(\hat{\bm\theta})}\le D$.}

Interestingly, if $\lambda_i^+( \ofim(\hat{\bm\theta}) ) < \frac{1}{\sigma_{\max}}\left(1-\frac{1}{N}\right)$,
the third term on the RHS of \cref{eq:og} becomes negative.
In the extreme case
when $\lambda_i^+( \ofim(\hat{\bm\theta}) )$ tends to zero,
$ \frac{1}{2}\log \left( \sigma_{\max} \lambda_i^+( \ofim(\hat{\bm\theta}) ) + \frac{1}{N} \right)
\to -\frac{1}{2}\log N
$,
which cancels out the model complexity penalty in the term $\frac{\rank{\ofim(\hat{\bm\theta})}}{2}\log{N}$.
In other words, the corresponding parameter is added free (without increasing the model complexity).
Informally, we call similar terms that are helpful in decreasing the complexity
while contributing to model flexibility the \emph{negative complexity}.

We have
\begin{align*}
      \sum_{i=1}^{\rank{\ofim(\hat{\bm\theta})}}
      \log
      \left( \sigma_{\min}
      \lambda_i^+(\ofim(\hat{\bm\theta}))
      + \frac{1}{N}
      \right)
      &\le
       \sum_{i=1}^{\rank{\ofim(\hat{\bm\theta})}}
      \log
      \left( \lambda_i^+(\ofim(\hat{\bm\theta})\diag{\bm\sigma})
      + \frac{1}{N}
      \right)\nonumber\\
      &\le
      \sum_{i=1}^{\rank{\ofim(\hat{\bm\theta})}}
      \log
      \left( \sigma_{\max}
      \lambda_i^+(\ofim(\hat{\bm\theta}))
      + \frac{1}{N}
      \right),
\end{align*}
where $\sigma_{\max}$ and $\sigma_{\min}$  denote the largest and smallest elements of $\bm\sigma$, respectively.
Therefore the term can be bounded based on the spectrum of $\ofim(\hat{\bm\theta})$.
\rev{If $\bm{\sigma}=\sigma \bm{1}$, where $\sigma>0$,
then both of the above ``$\le$''s become equalities.
In this case, we let $D\to\infty$ and rewrite the razor in terms of the spectrum density
$\rho_{\fim}(\lambda)$ of $\ofim(\hat{\bm\theta})$:
\begin{align}
\calO_G =
   &
    - \log{p}(\bm{X}\,\vert\,\hat{\bm\theta})
    + \frac{\rank{\ofim(\hat{\bm\theta})}}{2}
    \expect_{\rho_{\fim}(\lambda)}
    \log \left(N \sigma \lambda + 1 \right)
    + O(1).
\end{align}
Note $\rank{\ofim(\hat{\bm\theta})}=\hat{d}(\hat{\bm\theta})$ is the local dimensionality at $\hat{\bm\theta}$, which could have a smaller order than $D$, especially when $N$ is finite. If $\rho_\fim(\lambda)$ is highly concentrated around $0$ as shown in~\cite{PathologicalSpectrumFIM-2021}, then the expectation of $\log \left(N \sigma \lambda + 1 \right)$ can be roughly approximated as zero. This approximation is also linked to the low intrinsic complexity of DNNs.}

The Gaussian prior $p_G$ is helpful to give simple and intuitive expressions of $\calO_G$.
However, the problem in choosing $p_G$ is two fold. First, it is not invariant. Under a
reparametrization (\eg normalization or centering techniques), the Gaussian
prior in the new parameter system does not correspond to the original prior.
Second, it double counts equivalent models.  Because of the many singularities of the
neuromanifold, a small dynamic in the parameter system may not change the
prediction model. However, the Gaussian prior is defined in a real vector
space and may not fit in this singular semi-Riemannian structure.
Gaussian distributions are defined on Riemannian manifolds~\cite{said2017gaussian}
which lead to potential extensions of the discussed prior $p_G(\bm\theta)$.

\section{The Razor based on Jeffreys' Non-informative Prior}\label{sec:jeffreys}
Jeffreys' prior is specified by
$p_J(\bm\theta)\propto\sqrt{\vert\fim(\bm\theta)\vert}$. It is \emph{non-informative} in
the sense that no neural network model $\bm\theta_1$ is prioritized over any
other model $\bm\theta_2$.
It is invariant to the choice of the coordinate system.
Under a reparameterization $\bm\theta\to\bm\eta$,
\begin{align*}
\sqrt{\vert\fim(\bm\eta)\vert} \deta
&= \sqrt{ \left\vert
               \left(\frac{\partial\bm\theta}{\partial\bm\eta}\right)^\top
               \fim(\bm\theta)
               \frac{\partial\bm\theta}{\partial\bm\eta}
           \right\vert} \cdot \deta
=
\sqrt{\vert\fim(\bm\theta)\vert}
\cdot
\left(
\left\vert \frac{\partial\bm\theta}{\partial\bm\eta} \right\vert\deta
\right)
=
\sqrt{\vert\fim(\bm\theta)\vert} \dtheta, \nonumber
\end{align*}
showing that the Riemannian volume element is the same in different coordinate systems.
Unfortunately,
the Jeffreys' prior is \emph{not} well defined on the lightlike
neuromanifold $\calM$, where the metric $\fim(\bm\theta)$ is degenerate and $\sqrt{\vert\fim(\bm\theta)\vert}$ becomes zero.
The stratifold structure of $\calM$, where $d(\bm\theta)$
varying with $\bm\theta\in\calM$, makes it difficult to properly define
the base measure $d\bm\theta$ and integrate functions as in
\cref{eq:general}. From a mathematical standpoint, one has to integrate on the
screen distribution $\scr$, which has a Riemannian structure.  We refer the
reader to \cite{alphaprior-2005,weylprior} for other extensions of Jeffreys'
prior.

In this paper, we take a simple
approach by examining a submanifold of $\mathcal{M}$ denoted as $\widetilde{\mathcal{M}}$
and parameterized by
$\bm\xi$, which has a Riemannian metric $\fim(\bm\xi)\succ0$ that is induced by the FIM $\fim(\bm\theta)\succeq0$ and the mapping $\bm\xi\to\bm\theta$.
The dimensionality of $\widetilde{\calM}$
is upper-bounded by the local dimensionality $d(\bm\theta)$.
Intuitively, any infinitesimal dynamic on $\widetilde{\mathcal{M}}$ means such a change of neural network parameters that leads to a non-zero change of the global predictive model $\bm{z}\to{y}$.
\rev{For example, $\widetilde{\calM}$ can be defined based on a subset of sensitive parameters.
In theory, we would like to construct $\widetilde{\calM}$ so that it is representative of $\calM$, meaning that $\dim(\widetilde{\calM})$ is close to the local dimensionality $d(\bm\theta)$, and at the same time $\widetilde{\calM}$ remains Riemannian.}
The following results are constrained to the choice of the submanifold $\widetilde{\mathcal{M}}$.

In \cref{eq:general}, let $\kappa(\bm\xi)=\sqrt{\vert\fim(\bm\xi)\vert}$.
We further assume
\begin{description}
\item[(\text{\bf{A6}})]
$0<\int_{\widetilde{\calM}}
      \sqrt{\vert\fim(\bm\xi)\vert}
      \dxi<\infty$;
\end{description}
meaning that the Riemannian volume of $\widetilde{\calM}$ is bounded.
After straightforward derivations, we arrive at
\begin{align}\label{eq:oj}
\calO_J(\bm\xi)
=
&
- \log{p}(\bm{X}\,\vert\,\hat{\bm\xi})
+ \log\int_{\tildeM}
      \sqrt{\vert\fim(\bm\xi)\vert}
      \dxi
      \nonumber\\
&
- \log\int_{\widetilde{\calM}}
\exp\left(
-\frac{N}{2}(\bm\xi-\hat{\bm\xi})^\top
\ofim(\hat{\bm\xi})
(\bm\xi-\hat{\bm\xi})
\right)
\sqrt{\vert\fim(\bm\xi)\vert}
\dxi\nonumber\\
= &
- \log{p}(\bm{X}\,\vert\,\hat{\bm\xi})
+ \log\int_{\tildeM}
      \sqrt{\vert\fim(\bm\xi)\vert} \dxi
- \log\int_{\widetilde{\calM}}
    \omega(\bm\xi)
    \sqrt{\vert\fim(\bm\xi)\vert} \dxi,
\end{align}
where
$\omega(\bm\xi)\defeq\exp\left(
-\frac{N}{2}(\bm\xi-\hat{\bm\xi})^\top
\ofim(\hat{\bm\xi})
(\bm\xi-\hat{\bm\xi})
\right)$ is a shorthand.
Let us examine the meaning of $\calO_J(\bm\xi)$.
As $\fim(\bm\xi)$ is the Riemannian metric of $\tildeM$ based on information geometry,
$\sqrt{\vert\fim(\bm\xi)\vert} \dxi$ is a Riemannian volume element (volume form).
In the second term on the RHS of \cref{eq:oj}, the integral
$\int_{\widetilde{\calM}} \sqrt{\vert\fim(\bm\xi)\vert} \dxi$ is the
information volume, or the total ``number'' of different DNN models~\citep{mbCPD} on $\tildeM$.
In the last (third) term, because $0<\omega(\bm\xi)\le1$,
the integral on the LHS of
$$
\int_{\tildeM} \omega(\bm\xi)
\sqrt{\vert\fim(\bm\xi)\vert}
\dxi
\le
\int_{\tildeM}
\sqrt{\vert\fim(\bm\xi)\vert}
\dxi
$$
means a ``weighted volume'' of $\tildeM$, where the positive weights $\omega(\bm\xi)$ are determined by the observed FIM $\ofim(\hat{\bm\xi})$.
Combining these two terms, the model complexity is the log-ratio between the unweighted volume and the weighted volume and is lower bounded by 0.

Assume the spectrum decomposition
$\ofim(\hat{\bm\xi})=\bm{Q} \diag{\lambda_i^+(\ofim(\hat{\bm\xi}))} \bm{Q}^\top$,
where $\bm{Q}$ has orthonormal columns,
and $\lambda_i^+(\ofim(\hat{\bm\xi}))$
are the positive eigenvalues of
$\ofim(\hat{\bm\xi})$. \Cref{eq:oj} becomes
\begin{align}
\calO_J(\bm\zeta)
=
&
- \log{p}(\bm{X}\,\vert\,\hat{\bm\zeta})
+ \log\int_{\tildeM}
      \sqrt{\vert\fim(\bm\zeta)\vert}
      \dxi\nonumber\\
&
-\log\int_{\tildeM}
\exp\left(
-\frac{N}{2}
\sum_{i=1}^{\rank{\ofim(\hat{\bm\xi})}}
\lambda_i^+
\left(\ofim(\hat{\bm\xi})\right)
(\bm\zeta_i-\hat{\bm\zeta}_i)^2
\right)
\sqrt{\vert\fim(\bm\zeta)\vert}
\dzeta,\label{eq:ojzeta}
\end{align}
where $\bm\zeta=\bm{Q}^\top \bm\xi$ is an orthogonal transformation of $\bm\xi$, and $\calO_J$ is invariant to such transformations.
If an eigenvalue of $\ofim(\hat{\bm\xi})$ has an order of $o(\frac{1}{N})$,
the last two terms in \cref{eq:ojzeta} cancel out in the corresponding direction,
meaning no complexity is added.
This is similar to how the positive and negative complexity terms cancel out in \cref{eq:og} -- small eigenvalues of $\ofim(\hat{\bm\xi})$ are helpful to enhance the representation power of DNNs without increasing the model complexity.
Only eigenvalues that are large enough contribute significantly to the model complexity.

\rev{In the rest of this section, we connect our $\calO_J$ with previous formulations of MDL~\cite{bMDL,mbCPD}.
Observe that $\omega(\bm\xi)$ in \cref{eq:oj} resembles a Gaussian density up to a scaling factor.}
If $\ofim(\hat{\bm\xi})$ has full rank, we can further write
\begin{align}
\calO_J(\tildeM)
=
& - \log{p}(\bm{X}\,\vert\,\hat{\bm\xi})
+\frac{\dim(\widetilde{\calM})}{2} \log \frac{N}{2\pi}
+ \log\int_{\widetilde{\calM}}
      \sqrt{\vert\fim(\bm\xi)\vert}
      \dxi\nonumber\\
&
-\log\int_{\tildeM}
G\left( \bm{\xi} \mid \hat{\bm\xi}, \frac{1}{N}\ofim^{-1}(\hat{\bm\xi})
\right)
\frac{\vert\fim(\bm\xi)\vert^{1/2}}{\vert \ofim(\hat{\bm\xi})\vert^{1/2}}
\dxi.\label{eq:oj2}
\end{align}
By assumption (A6), the RHS of \cref{eq:oj} is well defined, while the RHS of \cref{eq:oj2} is only meaningful for a full rank $\ofim(\hat{\bm\xi})$.
If $\ofim(\hat{\bm\xi})$ is \emph{not} invertible, one can consider the limit case when the zero eigenvalues of $\ofim(\hat{\bm\xi})$ are replaced by a small $\epsilon>0$ and still apply the expression in \cref{eq:oj2}.
One has to note that
$$
\int_{\tildeM}
G\left( \bm{\xi} \mid \hat{\bm\xi}, \frac{1}{N}\ofim^{-1}(\hat{\bm\xi})
\right) \dxi \le 1,
$$
as the integral is over $\tildeM$ which is a subset of $\Re^{\dim(\tildeM)}$.
The last term on the RHS of \cref{eq:oj2} resembles an expectation \wrt a Gaussian distribution centered at $\hat{\bm\xi}$ on $\widetilde{\calM}$, except that the
Gaussian density may be truncated by $\widetilde{\calM}$.
One can therefore take the rough approximation based on the mean of the Gaussian:
\begin{equation}\label{eq:naturalness}
-\log
\int_{\tildeM}
G\left( \bm{\xi} \mid \hat{\bm\xi}, \frac{1}{N}\ofim^{-1}(\hat{\bm\xi})
\right)
\frac{\vert\fim(\bm\xi)\vert^{1/2}}{\vert \ofim(\hat{\bm\xi})\vert^{1/2}}
\dxi
\approx
\frac{1}{2} \log
\frac{\vert \ofim(\hat{\bm\xi})\vert}{\vert \fim(\hat{\bm\xi})\vert}.
\end{equation}
\rev{Under this approximation,
\cref{eq:oj2} gives the MDL criterion discussed in~\cite{bMDL,mbCPD},
where the term on the RHS of~\cref{eq:naturalness} is interpreted as a penalty to models that lack robustness and are sensitive to the choice of parameters.}
We therefore consider the spectrum of both matrices $\fim(\bm\xi)$ and $\ofim(\bm\xi)$, noting that in the large sample limit $N\to\infty$,
they become identical.
Because of the finite $N$, the observed FIM $\ofim(\hat{\bm\xi})$ is singular in potentially many directions.
\rev{The true FIM $\fim(\hat{\bm\xi})$ can be regarded as the sum of the observed FIM $\ofim({\hat{\bm\xi}})$ and the FIM \wrt unobserved samples, up to a scaling factor.
Based on how $\widetilde{\calM}$ is constructed, $\fim(\hat{\bm\xi})\succ0$ is positive definite and
suffers less from singularities.
In the directions where $\ofim(\hat{\bm\xi})$ is nearly singular, the log-ratio
$\log {\vert \ofim(\hat{\bm\xi})\vert}/{\vert \fim(\hat{\bm\xi})\vert}$ contributes significantly and negatively to the model complexity.
As a result, \cref{eq:naturalness} serves as a negative complexity term and explains how singularities of $\ofim(\hat{\bm\xi})$ correspond to the simplicity of DNNs.}

Compared with $\calO_G$, $\calO_J$ is based on a more accurate geometric modeling, However, it is hard to be computed numerically, \rev{as it depends on how $\widetilde{\calM}$ is constructed, and $\fim(\bm\xi)$ and $p(\bm{z})$ which are unknown due to limited observations.}
Despite that $\calO_G$ and $\calO_J$ have different expressions, their preference to model dimensions with
small Fisher information (as in DNNs) is similar.

Hence, we can conclude that
\emph{the intrinsic complexity of a DNN is affected by the singularity and spectral properties
of the Fisher information matrix.}

\section{Related Work}\label{sec:related}

The dynamics of supervised learning of a DNN describes a trajectory on
the parameter space of the DNN geometrically modeled as a manifold when
endowed with the FIM
  (e.g., ordinary/natural gradient descent learning the parameters of a
MLP).
Singular regions of the neuromanifold~\citep{wzDOL} correspond to non-identifiable
parameters with rank-deficient FIM, and the learning trajectory
typically exhibits chaotic patterns~\citep{aoDOL} with the singularities which
translate into slowdown plateau phenomena when plotting the loss
function value against time.
By building an elementary singular DNN, \citep{aoDOL} (and references therein) showed
that \rev{stochastic gradient descent}
learning dynamics yields a Milnor-type attractor with both
attractor/repulser subregions where the learning trajectory is attracted
in the attractor region, then stay a long time there before
escaping through the repulser region.
The natural gradient is shown to be free of critical slowdowns.
Furthermore, although DNNs have potentially many singular regions,
the interaction of elementary units cancels out the
Milnor-type attractors.
It was shown~\citep{opSCE} that skip connections are helpful to reduce the effect of singularities.
However, a full understanding of the learning
dynamics~\citep{ykSMA} for generic DNN architectures with multiple
output values or recurrent DNNs is yet to be investigated.

The MDL criterion has undergone several fundamental revisions,
\rev{such as the original crude MDL~\citep{rMBS} and refined MDL with the introduction of stochastic complexity~\citep{rFIA,brTMD},
and the NML~\cite{shtarkov87,nml} as a modern refinement.}
We refer the reader to the book \citep{pTMDL} for a comprehensive introduction
to this area and \cite{grMDL} for a recent review.
\rev{We should also mention that the relationship between MDL and generalization
has been explored in the PAC-MDL framework~\cite{MDLbarroncover,MDLzhang,pTMDL,pmlr-v98-grunwald19a,PACMDLbounds}.
See \cite{grMDL} (section 6.4) for related remarks.
}

\rev{The relationship between MDL and information geometry is well established~\cite{rFIA,aIGA,bMDL,mbCPD,murata1994NIC}. For example, they both rely on fundamental concepts such as the Fisher information.
The geometric complexity of statistical models is commonly formulated using tools from information geometry~\cite{aANL,rFIA,bMDL,murata1994NIC}.
The stochastic complexity in singular mixture models can be bounded~\cite{yamazaki2003singularities} and therefore is smaller than that of regular models.
On this line of research,} our derivations based on a Taylor expansion of the log-likelihood are similar to
\cite{bMDL}. This technique is also used for deriving natural gradient optimization for deep learning~\citep{pbRNG,aoDOL,lin2023simplifying}.

Recently, MDL has been ported to deep learning~\citep{boTDL} focusing on
variational methods. Practical techniques such as weight sharing~\cite{wann},
binarization~\cite{bnn}, model compression~\cite{compress}, etc.,
follow similar principles of MDL.
\rev{In the same community, many efforts have been made to develop a theory of deep learning}, for example, based on PAC-Bayes theory~\citep{nbEGI},
statistical learning theory~\citep{zbUDL}, algorithmic information
theory~\citep{plDLG}, information geometry~\citep{lpFRM}, geometry of the DNN
mapping~\citep{rpOTE}, or through defining an intrinsic
dimensionality~\citep{lfMTI} that is much smaller than the network size.
Our analysis depends on $\ofim(\hat{\bm\theta})$ and therefore is related
to the flatness/sharpness of the local minima~\citep{flat,dpSMC},
which is known to affect generalization.
Using advanced mathematical tools such as random matrix theory,
investigations are conducted on the spectrum of the input-output Jacobian
matrix~\citep{psTEO}, the Hessian matrix \wrt the neural network
weights~\citep{pbGON}, and the
FIM~\citep{kaUSO,pwTSO,SpectrumFIM-2021,PathologicalSpectrumFIM-2021,papyan}.

\section{Conclusion}\label{sec:con}

We consider mathematical tools from singular semi-Riemannian geometry
to study the locally varying intrinsic dimensionality of a deep learning model.
These models fall in the category of non-identifiable parameterizations.
We take a meaningful step to quantify \emph{geometric singularity}
through the notion of local dimensionality $d(\bm\theta)$ yielding a singular semi-Riemannian neuromanifold with varying metric signature.
We show that  $d(\bm\theta)$ grows at most linearly with the sample size $N$.
Recent findings show that the spectrum of the Fisher information matrix shifts
towards $0^{+}$ with a large number of small eigenvalues.
We show that these singular dimensions help to reduce the model complexity.
As a result, we contribute a simple and general MDL for deep learning.
It provides theoretical insights on the description length of DNNs.
DNNs benefit from a high-dimensional parameter space in that the singular
dimensions impose a negative complexity to describe the data,
which can be seen in our derivations based on Gaussian and Jeffreys' priors.
\rev{How the short description length is connected to the empirical performance of DNNs and related generalization bounds require further examinations. This is not addressed in the current work.}
A more careful analysis of the FIM's spectrum, e.g. through considering higher-order
terms, could give more practical formulations of the proposed criterion.
We leave empirical studies as potential future work.
 
\begin{appendices}
    \section{Proof of \texorpdfstring{$\ofim(\hat{\bm\theta})=\hat{\fim}(\hat{\bm\theta})$}{J(theta)=I(theta)}}

\begin{proof}
\begin{align*}
p(y_i\,\vert\,\bm{z}_i,\bm\theta)
=
\exp\left(
    \onehot(y_i)^\top \bm{h}^L(\bm{z}_i)
    -
    \log\sum_{j}\exp(h^L_j(\bm{z}_i))
\right),
\end{align*}
where $\onehot(y)$ is the binary vector with the same dimensionality as
$\bm{h}^L(\bm{z}_i)$, with the $y$'th bit set to 1 and the rest bits set
to 0. Therefore,
\begin{align*}
\frac{\partial \log p(y_i\,\vert\,\bm{z}_i,\bm\theta)}
{\partial\bm\theta}
=
\left[ \frac{\partial\bm{h}^L}{\partial\bm\theta} \right]^\top
\left[ \onehot(y_i) - \softmax(\bm{h}^L(\bm{z}_i)) \right].
\end{align*}
Therefore,
\begin{align}\label{eq:hess}
\frac{\partial^2\log p(y_i\,\vert\,\bm{z}_i,\bm\theta)}{\partial\bm\theta\partial\bm\theta^\top}
=
& \sum_j \left[ \onehot(y_i) - \softmax(\bm{h}^L(\bm{z}_i)) \right]_j
\frac{\partial^2\bm{h}^L_j}{\partial\bm\theta\partial\bm\theta^\top}
- \left[ \frac{\partial\bm{h}^L}{\partial\bm\theta} \right]^\top
\cdot \bm{C}_i \cdot
\frac{\partial\bm{h}^L}{\partial\bm\theta}.
\end{align}
where
\begin{align*}
\bm{C}_i
=
\frac{\partial\softmax(\bm{h}^L(\bm{z}_i))}{\partial\bm{h}^L(\bm{z}_i)}
=
\diag{\bm{o}_i} -\bm{o}_i \bm{o}_i^\top,
\quad
\bm{o}_i
= \softmax(\bm{h}^L(\bm{z}_i)).
\end{align*}
By (\textbf{A1}), at the MLE $\hat{\bm\theta}$,
\begin{equation*}
\forall{i},\quad
\softmax(\bm{h}^L(\bm{z}_i)) = \onehot(y_i).
\end{equation*}
Therefore
\begin{equation*}
\forall{i},
\quad
-\frac{\partial^2\log p(y_i\,\vert\,\bm{z}_i,\bm\theta)}{\partial\bm\theta\partial\bm\theta^\top}
=
\left[ \frac{\partial\bm{h}^L}{\partial\bm\theta} \right]^\top
\cdot \bm{C}_i \cdot
\frac{\partial\bm{h}^L}{\partial\bm\theta}.
\end{equation*}
Taking the sample average on both sides, we get
\begin{equation*}
\ofim(\hat{\bm\theta}) = \hat{\fim}(\hat{\bm\theta}).
\end{equation*}
\end{proof}

\section{Proof of \Cref{thm:kernel}}\label{sec:prooflemma}

\begin{proof}
    If $(\bm\theta, \sum_{j}\alpha_j\partial\theta_j)\in\rad$,
    Then
    \begin{align*}
        \left\langle\sum_{j}\alpha_j\partial\theta_j,
                \sum_{j}\alpha_j\partial\theta_j\right\rangle_{\fim(\bm\theta)} = 0.
    \end{align*}
    In matrix form, it is simply $\bm\alpha^\top \fim(\bm\theta) \bm\alpha=0$.
    We have the analytical expression
    \begin{align*}
    \fim(\bm\theta)
    &= \expect_p
    \left[\left(
    \frac{\partial\bm{h}^L(\bm{z})}{\partial\bm\theta} \right)^\top
    \bm{C}(\bm{z})
    \frac{\partial\bm{h}^L(\bm{z})}{\partial\bm\theta}
    \right].
    \end{align*}
    Therefore
    \begin{align*}
    E_p
    \left[
    \left( \frac{\partial\bm{h}^L(\bm{z})}{\partial\bm\theta}\bm\alpha \right)^\top
    \bm{C}(\bm{z}) \frac{\partial\bm{h}^L(\bm{z})}{\partial\bm\theta}
    \bm\alpha
    \right]
    =0.
    \end{align*}
    By noting that $\bm{C}(\bm{z})\succeq0$ is psd, we have almost surely that
    \begin{align*}
    \left( \frac{\partial\bm{h}^L(\bm{z})}{\partial\bm\theta}\bm\alpha \right)^\top
    \bm{C}(\bm{z}) \frac{\partial\bm{h}^L(\bm{z})}{\partial\bm\theta}\bm\alpha = 0.
    \end{align*}
    Any eigenvector of $\bm{C}(\bm{z})$ associated with the zero eigenvalues must be
    a multiple of $\bm{1}$.
    Indeed,
    \begin{equation*}
        \bm{v}^\top \bm{C}(\bm{z}) \bm{v}
        =
        \bm{v}^\top
        \left(
        \diag{\bm{o}(\bm{z})}-\bm{o}(\bm{z})\bm{o}(\bm{z})^\top \right)
        \bm{v}
        =
        \sum_j o_{j}(\bm{z}) (v_j-\sum_{j}o_{j}(\bm{z}) v_j)^2
        =
        0
        \Leftrightarrow
        \bm{v} \propto \bm{1},
    \end{equation*}
    where $o_{j}(\bm{z})>0$ is the $j$'th element of $\bm{o}(\bm{z})$. Hence,
    almost surely
    \begin{align*}
    \frac{\partial\bm{h}^L(\bm{z})}{\partial\bm\theta}\bm\alpha = \lambda(\bm{z}) \bm{1}.
    \end{align*}
\end{proof}

\emph{Remark.}
$\bm\alpha$ is associated with a tangent vector in $\rad$, meaning a dynamic
along the lightlike dimensions.
The Jacobian $\frac{\partial\bm{h}^L(\bm{z})}{\partial\bm\theta}$
is the local linear approximation of the mapping $\bm\theta\to\bm{h}^L(\bm{z})$.
By \cref{thm:kernel}, with probability 1 such a dynamic leads to uniform increments in the output
units, meaning
$\bm{h}^L(\bm{z})\to\bm{h}^L(\bm{z})+\lambda(\bm{z})\bm{1}$, $\forall{i}$, and therefore the
output distribution $\softmax\left(\bm{h}^L(\bm{z})\right)$ is not affected.
In summary, we have verified that the radical distribution does not affect the
neural network mapping.

\section{Proof of \Cref{thm:dim}}

\begin{proof}
\begin{align*}
\hat{d}(\bm\theta)
&
= \rank{\hat{\fim}(\bm\theta)}
= \rank{
    \sum_{i=1}^N
    \left(\frac{\partial \bm{h}^L(\bm{z}_i)}{\partial\bm\theta}\right)^\top
    \bm{C}_i
    \frac{\partial \bm{h}^L(\bm{z}_i)}{\partial\bm\theta}
}
\nonumber\\
& \le
\sum_{i=1}^N \rank{
\left(\frac{\partial \bm{h}^L(\bm{z}_i)}{\partial\bm\theta}\right)^\top
\bm{C}_i
\frac{\partial \bm{h}^L(\bm{z}_i)}{\partial\bm\theta}
}
\le
\sum_{i=1}^N
\min\left\{
\rank{\frac{\partial \bm{h}^L(\bm{z}_i)}{\partial\bm\theta}},
\rank{\bm{C}_i} \right\}.
\end{align*}

Note the matrix $\frac{\partial \bm{h}^L(\bm{z}_i)}{\partial\bm\theta}$
has size $m\times{}D$, and $\bm{C}_i$ has size $m\times{m}$ and rank $(m-1)$.
We also have $\hat{d}(\bm\theta)=\rank{\hat{\fim}(\bm\theta)}\le{D}=\dim(\bm\theta)$.
Therefore
\begin{equation*}
\hat{d}(\bm\theta) \le
\sum_{i=1}^N
\min\left\{
\rank{\frac{\partial \bm{h}^L(\bm{z}_i)}{\partial\bm\theta}},
m-1 \right\}.
\end{equation*}
\end{proof}

\section{Proof of \Cref{thm:dim2}}
\begin{proof}
We only prove the upper bound of $d(\bm\theta)$.
The upper bound of $\hat{d}(\bm\theta)$ can be proved similarly.

\begin{align*}
d(\bm\theta)
&
= \rank{\fim(\bm\theta)}
= \rank{
\expect_{p(\bm{z})}
\left[
    \left(\frac{\partial \bm{h}^L(\bm{z})}{\partial\bm\theta}\right)^\top
    \bm{C}(\bm{z})
    \frac{\partial \bm{h}^L(\bm{z})}{\partial\bm\theta}
    \right]
}.
\end{align*}
That means $d(\bm\theta)$ is the dimensionality of the image of
$\expect_{p(\bm{z})}
\left[
    \left(\frac{\partial \bm{h}^L(\bm{z})}{\partial\bm\theta}\right)^\top
    \bm{C}(\bm{z})
    \frac{\partial \bm{h}^L(\bm{z})}{\partial\bm\theta}
    \right]$.

$\forall\bm{\theta}$, we have
\begin{align*}
\expect_{p(\bm{z})}
\left[
    \left(\frac{\partial \bm{h}^L(\bm{z})}{\partial\bm\theta}\right)^\top
    \bm{C}(\bm{z})
    \frac{\partial \bm{h}^L(\bm{z})}{\partial\bm\theta}
    \right]\bm\theta
&=
\expect_{p(\bm{z})}
\left[
    \left(\frac{\partial \bm{h}^L(\bm{z})}{\partial\bm\theta}\right)^\top
    \bm{C}(\bm{z})
    \frac{\partial \bm{h}^L(\bm{z})}{\partial\bm\theta} \bm\theta
    \right]\nonumber\\
&=
\expect_{p(\bm{z})}
\left[
    \left(\frac{\partial \bm{h}^L(\bm{z})}{\partial\bm\theta}\right)^\top
    \bm{\beta}(\bm{z})
\right],
\end{align*}
where $\bm\beta(\bm{z})= \bm{C}(\bm{z})
    \frac{\partial \bm{h}^L(\bm{z})}{\partial\bm\theta} \bm\theta$ is an $m$-dimensional vector.
Therefore
\begin{align*}
    \expect_{p(\bm{z})}
\left[
    \left(\frac{\partial \bm{h}^L(\bm{z})}{\partial\bm\theta}\right)^\top
    \bm{C}(\bm{z})
    \frac{\partial \bm{h}^L(\bm{z})}{\partial\bm\theta}
    \right]\bm\theta
\in
\mathrm{span}\;
\bigcup_{\bm{z}\in\mathrm{supp}(p)}
\mathrm{Row}
 \left(\frac{\partial \bm{h}^L(\bm{z})}{\partial\bm\theta}\right).
\end{align*}
Letting $\bm\theta$ vary in $\Re^{D}$, and applying $\dim(\cdot)$ on both sides, the statement follows immediately.
\end{proof}

\section{Proof of \Cref{thm:dim3}}
\begin{proof}
We have the total derivative
\begin{align*}
\de \bm{h}^L(\bm{z})
=
\sum_{l=1}^L
\bm{W}^L \bm{\Phi}^{L-1} \bm{W}^{L-1} \cdots
\bm{\Phi}^{l}
\left(
\de \bm{W}^l \bm{z}^{l-1}
+ \de \bm{b}^l
\right).
\end{align*}

Therefore,
\begin{equation*}
\forall{\bm{w}}\in\Re^{\dim(\bm{w}^l)},
\quad
\frac{\partial\bm{h}^L(\bm{z})}{\partial\bm{w}^l}
\bm{w}
=
\bm{W}^L
\bm{\Phi}^{L-1} \bm{W}^{L-1} \cdots
\bm{\Phi}^l
\mathrm{mat}(\bm{w})
\begin{pmatrix}
\bm{z}^{l-1}\\
1
\end{pmatrix},
\end{equation*}
where $\mathrm{mat}(\cdot)$ means to rearrange the vector into a matrix. Therefore,
\begin{equation*}
\rank{\frac{\partial\bm{h}^L(\bm{z})}{\partial\bm{w}^l}}
\le
\rank{\bm{W}^L \bm{\Phi}^{L-1} \bm{W}^{L-1} \cdots \bm{\Phi}^l}.
\end{equation*}
The second ``$\le$'' in the statement is because
\begin{align*}
\rank{\bm{W}^L \bm{\Phi}^{L-1} \bm{W}^{L-1} \cdots \bm{\Phi}^l}
&\le
\min\left\{
\rank{\bm{W}^L},
\rank{\bm{\Phi}^{L-1}},
\cdots
\rank{\bm{\Phi}^l}
\right\}\nonumber\\
&\le
\min\left\{
\rank{\bm{\Phi}^{L-1}},
\cdots,
\rank{\bm{\Phi}^l}
\right\}\nonumber\\
&=
\min_{s=l}^{L-1} \rank{\bm{\Phi}^s}.
\end{align*}
\end{proof}

\section{Metric Signature of the Neuromanifold}

The metric signature of $\calM$
\begin{equation*}
\left( d(\bm\theta), 0, D-d(\bm\theta) \right)
\end{equation*}
is straightforward from the fact that $\fim(\bm\theta)$ is
positive semi-definite (there is no negative eigenvalues),
and the local dimensionality $d(\bm\theta)$, by
definition, is $\rank{\fim(\bm\theta)}$
(the number of non-zero eigenvalues).

We also show that $\rank{\ofim(\bm\theta)} \neq \hat{d}(\bm\theta)$.
Recall that
    $\hat{d}(\bm\theta) = \rank{\hat{\fim}(\bm\theta)}$, and
    \begin{equation*}
    \rank{\ofim(\bm\theta)}
    =
    \rank{\frac{\partial^2\ell}{\partial\bm\theta\bm\theta^{\top}}}
    =
    \rank{\sum_{i}\frac{\partial^2\ell_i}{\partial\bm\theta\bm\theta^{\top}}},
    \end{equation*}
    where $\ell$ is the log-likelihood, and $\ell_i=\log{ p(y_i\,\vert\,\bm{z}_i,\bm\theta) }$.
    We write the analytical form of the elementwise Hessian
    \begin{align*}
        \frac{\partial^2\ell_i}{\partial\bm\theta\partial\bm\theta^\top}
        =
        \sum_{j=1}^m
        \frac{\partial h_j^L(\bm{z}_i)}{\partial\bm\theta\partial\bm\theta^\top}
        (\mathrm{OneHot}_j(y) - \softmax_j(\bm{h}^L)) - \fim(\bm\theta),
    \end{align*}
    where $\mathrm{OneHot}(\cdot)$ denote the one-hot vector associated with the given
    target label $y$. Therefore
    \begin{align*}
        \bm\alpha^\top
        \frac{\partial^2\ell_i}{\partial\bm\theta\partial\bm\theta^\top}
        \bm\alpha
        &=
        \sum_{j=1}^m
        \bm\alpha^\top
        \left(
        \frac{\partial h_j^L(\bm{z}_i)}{\partial\bm\theta\partial\bm\theta^\top}
        \bm\alpha
        \right)
        (\mathrm{OneHot}_j(y) - \softmax_j(\bm{h}^L))
        -
        \bm\alpha^\top
        \fim(\bm\theta)
        \bm\alpha.
    \end{align*}
    Because of the first term on the RHS, the kernels of the two matrices
    $\ofim(\bm\theta)$ and $\hat{\fim}(\bm\theta)$ are different, and
    thus their ranks are also different.

\section{Proof of \Cref{thm:general}}

\begin{proof}
    As $\hat{\bm\theta}$ is the MLE, we have $\ofim(\hat{\bm\theta}) \succeq0$, and $\forall\bm\theta\in\calM$,
     \begin{equation*}
         -\frac{N}{2} (\bm\theta-\hat{\bm\theta})^\top
     \ofim(\hat{\bm\theta}) (\bm\theta-\hat{\bm\theta}) \le 0.
 \end{equation*}
 Hence,
 \begin{equation*}
     \expect_p \exp\bigg(
     -\frac{N}{2} (\bm\theta-\hat{\bm\theta})^\top
     \ofim(\hat{\bm\theta}) (\bm\theta-\hat{\bm\theta}) \bigg) \le 1.
 \end{equation*}
 Hence,
 \begin{equation*}
     - \log \expect_p \exp\bigg(
     -\frac{N}{2} (\bm\theta-\hat{\bm\theta})^\top
     \ofim(\hat{\bm\theta}) (\bm\theta-\hat{\bm\theta}) \bigg) \ge 0.
 \end{equation*}
 This proves the first ``$\le$''.

As $-\log(x)$ is convex, by Jensen's inequality, we get
\begin{align*}
&- \log \expect_p \exp\bigg(
     -\frac{N}{2} (\bm\theta-\hat{\bm\theta})^\top
     \ofim(\hat{\bm\theta}) (\bm\theta-\hat{\bm\theta}) \bigg)
     \nonumber\\
&\le
     \expect_p\bigg( - \log \exp \bigg(
     -\frac{N}{2} (\bm\theta-\hat{\bm\theta})^\top
     \ofim(\hat{\bm\theta}) (\bm\theta-\hat{\bm\theta}) \bigg)
     \bigg)\nonumber\\
&=
     \expect_p\bigg( \frac{N}{2} (\bm\theta-\hat{\bm\theta})^\top
     \ofim(\hat{\bm\theta}) (\bm\theta-\hat{\bm\theta}) \bigg)\nonumber\\
&=
     \frac{N}{2}
         \trace\left(
            \expect_p\bigg(
            \ofim(\hat{\bm\theta}) (\bm\theta-\hat{\bm\theta}) (\bm\theta-\hat{\bm\theta})^\top
            \bigg)
         \right)
     \nonumber\\
     &=
     \frac{N}{2}
         \trace\left(
             \ofim(\hat{\bm\theta})
             \left(
             (\mu(\bm\theta)-\hat{\bm\theta})
             (\mu(\bm\theta)-\hat{\bm\theta})^\top
             +\cov{\bm\theta} \right)
         \right).
\end{align*}
This proves the second ``$\le$''.
\end{proof}

\section{Proof of \Cref{thm:b}}

\begin{proof}
Due to the convexity of $-\log t$, we have
\begin{align*}
\overline{
\{ M_f(\bm{t}_{:,1}), \cdots, M_f(\bm{t}_{:,m}) \} }
&=
\frac{1}{m} \sum_{j=1}^m
\left[ -\log
\left( \frac{1}{n}\sum_{i=1}^n \exp(-t_{ij}) \right)
\right]\nonumber\\
&\ge
-\log
\left[
\frac{1}{m} \sum_{j=1}^m
\frac{1}{n}\sum_{i=1}^n \exp(-t_{ij})
\right]
= M_f(\bm{T}).
\end{align*}
This proves the first ``$\le$''.
To prove the second ``$\le$'', we note that $-\log \frac{1}{n}\sum_{i=1}^n \exp(-t_i)$ is a concave function. Therefore
\begin{align*}
&\overline{ \{ M_f(\bm{t}_{:,1}), \cdots, M_f(\bm{t}_{:,m}) \} }
=
\frac{1}{m} \sum_{j=1}^m
\left[ -\log
\left( \frac{1}{n}\sum_{i=1}^n \exp(-t_{ij}) \right)
\right]\nonumber\\
\le
&
-\log
\left( \frac{1}{n}\sum_{i=1}^n \exp\left(-
\frac{1}{m} \sum_{j=1}^m t_{ij}\right) \right)
=
M_f\left( \{ \overline{\bm{t}_1}, \cdots, \overline{\bm{t}_n} \}\right).
\end{align*}
The last ``$\le$'' is based on the convexity of $-\log t$.
Once again, by Jensen's inequality, we have
\begin{align*}
M_f\left( \{ \overline{\bm{t}_1}, \cdots, \overline{\bm{t}_n} \}\right)
\le
\frac{1}{n}\sum_{i=1}^n
-\log\left( \exp\left(-
\frac{1}{m} \sum_{j=1}^m t_{ij}\right) \right)
=
\overline{\bm{T}}.
\end{align*}

\end{proof}

\section{Derivations of $\calO_G$}\label{sec:proofg}

We recall the general formulation in \cref{eq:general}:
\begin{align*}
\calO
\defeq
    & - \log{p}(\bm{X}\,\vert\,\hat{\bm\theta})
      + \log\int_{\calM}\kappa(\bm\theta)\dtheta \nonumber\\
    & - \log\int_{\calM}
        \kappa(\bm\theta) \exp\left(
        -\frac{N}{2} (\bm\theta-\hat{\bm\theta})^\top
        \ofim(\hat{\bm\theta}) (\bm\theta-\hat{\bm\theta})
    \right) \dtheta.
\end{align*}
If
$\kappa(\bm\theta) = \exp\left(-\frac{1}{2}\bm\theta^\top \diag{\frac{1}{\bm\sigma}} \bm\theta\right)$,
then the second term on the RHS is
\begin{align*}
&\log\int_{\calM}\kappa(\bm\theta)\dtheta
=
\log \int_{\calM}
\exp\left( -\frac{1}{2} \bm\theta^\top \diag{\frac{1}{\bm\sigma}} \bm\theta\right) \dtheta
\nonumber\\
&=
\frac{D}{2}\log2\pi
+
\frac{1}{2} \log\vert\diag{\bm\sigma}\vert
\nonumber\\ &
+
\log \int_{\calM}
\exp\left(-\frac{D}{2}\log{2\pi}
    - \frac{1}{2}\log\vert\diag{\bm\sigma}\vert
    - \frac{1}{2}\bm\theta^\top \diag{\frac{1}{\bm\sigma}} \bm\theta\right) \dtheta
\nonumber\\
&=
\frac{D}{2}\log2\pi
+ \frac{1}{2} \log\vert\diag{\bm\sigma}\vert
+ \log 1
=
\frac{D}{2}\log2\pi +
\frac{1}{2} \log\vert\diag{\bm\sigma}\vert.
\end{align*}
The third (last) term on the RHS is
\begin{align*}
    &
    - \log\int_{\calM}
        \kappa(\bm\theta) \exp\left(
        -\frac{N}{2} (\bm\theta-\hat{\bm\theta})^\top
        \ofim(\hat{\bm\theta}) (\bm\theta-\hat{\bm\theta})
    \right) \dtheta
    \nonumber\\
= &
    - \log\int_{\calM}
    \exp\left( -\frac{1}{2}\bm\theta^\top \diag{\frac{1}{\bm\sigma}} \bm\theta
    -\frac{N}{2} (\bm\theta-\hat{\bm\theta})^\top
        \ofim(\hat{\bm\theta}) (\bm\theta-\hat{\bm\theta})
    \right) \dtheta
    \nonumber\\
= &
    - \log\int_{\calM}
    \exp\left(
        - \frac{1}{2} \bm\theta^\top \bm{A} \bm\theta
        + \bm{b}^\top \bm\theta + c
    \right) \dtheta,
\end{align*}
where
\begin{align*}
\bm{A}
=
N \ofim(\hat{\bm\theta}) + \diag{\frac{1}{\bm\sigma}}\succ0,
\quad
\bm{b}
=
N\ofim(\hat{\bm\theta})\hat{\bm\theta},
\quad
c
=
- \frac{N}{2} \hat{\bm\theta}^\top \ofim(\hat{\bm\theta})\hat{\bm\theta}.
\end{align*}
Then,
\begin{align*}
    &
    - \log\int_{\calM}
        \kappa(\bm\theta) \exp\left(
        -\frac{N}{2} (\bm\theta-\hat{\bm\theta})^\top
        \ofim(\hat{\bm\theta}) (\bm\theta-\hat{\bm\theta})
    \right) \dtheta
    \nonumber\\
=
    &
    - \log\int_{\calM}
    \exp\left(
        - \frac{1}{2} (\bm\theta-\bar{\bm\theta})^\top
        \bm{A} (\bm\theta-\bar{\bm\theta})
        + c + \frac{1}{2}\bar{\bm\theta}^\top \bm{A} \bar{\bm\theta}
    \right) \dtheta
    \nonumber\\
=
    &
    - \frac{D}{2}\log2\pi
    + \frac{1}{2}\log\vert\bm{A}\vert
    - c - \frac{1}{2}\bar{\bm\theta}^\top \bm{A} \bar{\bm\theta}
    \nonumber\\
    &
    - \log\int_{\calM}
    \exp\left(
        - \frac{D}{2}\log2\pi
        + \frac{1}{2}\log\vert\bm{A}\vert
        - \frac{1}{2} (\bm\theta-\bar{\bm\theta})^\top
        \bm{A} (\bm\theta-\bar{\bm\theta})
    \right) \dtheta\nonumber\\
=
    &
    - \frac{D}{2}\log2\pi + \frac{1}{2}\log\vert\bm{A}\vert
    - c
    - \frac{1}{2}\bar{\bm\theta}^\top \bm{A} \bar{\bm\theta},
\end{align*}
where $\bm{A}\bar{\bm\theta}=\bm{b}$.
To sum up,
\begin{align*}
    &\calO_G
    =
    - \log{p}(\bm{X}\,\vert\,\hat{\bm\theta})
      + \frac{D}{2}\log2\pi + \frac{1}{2} \log\vert\diag{\bm\sigma}\vert\nonumber\\
    & \quad\quad - \frac{D}{2}\log2\pi + \frac{1}{2} \log\vert\bm{A}\vert
      - c
      - \frac{1}{2}\bar{\bm\theta}^\top \bm{A} \bar{\bm\theta}\nonumber\\
    =
    & - \log{p}(\bm{X}\,\vert\,\hat{\bm\theta})
      + \frac{1}{2} \log\vert\diag{\bm\sigma}\vert
      + \frac{1}{2} \log\vert\bm{A}\vert
      - c
      - \frac{1}{2}\bar{\bm\theta}^\top \bm{A} \bar{\bm\theta},\nonumber\\
    =
    & - \log{p}(\bm{X}\,\vert\,\hat{\bm\theta})
      + \frac{1}{2} \log\vert\diag{\bm\sigma}\vert
      + \frac{1}{2} \log \vert N \ofim(\hat{\bm\theta}) + \diag{\frac{1}{\bm\sigma}} \vert
      \nonumber\\
    &  + \frac{N}{2} \hat{\bm\theta}^\top \ofim(\hat{\bm\theta})\hat{\bm\theta}
       -\frac{1}{2} \left(N\ofim(\hat{\bm\theta})\hat{\bm\theta}\right)^\top
        \left( N\ofim(\hat{\bm\theta}) + \diag{\frac{1}{\bm\sigma}} \right)^{-1}
        N\ofim(\hat{\bm\theta})\hat{\bm\theta}
        \nonumber\\
    =
    & - \log{p}(\bm{X}\,\vert\,\hat{\bm\theta})
      + \frac{1}{2}
        \log\vert N \ofim(\hat{\bm\theta})\diag{\bm\sigma} + \bm{I} \vert
      \nonumber\\
    & + \frac{1}{2}
        \hat{\bm\theta}^\top
        \left(\ofim(\hat{\bm\theta})\right)^\top
        \left(\ofim(\hat{\bm\theta}) + \frac{1}{N}\diag{\frac{1}{\bm\sigma}} \right)^{-1}
        \diag{\frac{1}{\bm\sigma}} \hat{\bm\theta}\nonumber\\
=&  - \log{p}(\bm{X}\,\vert\,\hat{\bm\theta})
      + \frac{1}{2}
        \log\vert N \ofim(\hat{\bm\theta})\diag{\bm\sigma} + \bm{I} \vert
+ \frac{1}{2}
        \hat{\bm\theta}^\top
        \ofim(\hat{\bm\theta})
        \left(
        \diag{\bm\sigma}
        \ofim(\hat{\bm\theta})
        + \frac{1}{N}\bm{I} \right)^{-1}
        \hat{\bm\theta}.
        \end{align*}
The last term does not scale with $N$ and has a smaller order as compared to other terms.
Indeed,
\begin{equation*}
\lim_{N\to\infty}
\ofim(\hat{\bm\theta})
\left(\ofim(\hat{\bm\theta}) + \frac{1}{N}\diag{\frac{1}{\bm\sigma}} \right)^{-1}
=
\ofim(\hat{\bm\theta}) \ofim(\hat{\bm\theta})^+,
\end{equation*}
where $\ofim(\hat{\bm\theta})^+$ is the Moore-Penrose inverse of $\ofim(\hat{\bm\theta})$. Hence, as $N\to\infty$,
\begin{align*}
 \frac{1}{2}
 \hat{\bm\theta}^\top
 \ofim(\hat{\bm\theta})
 \left(
 \diag{\bm\sigma}
 \ofim(\hat{\bm\theta})
 + \frac{1}{N}\bm{I} \right)^{-1}
 \hat{\bm\theta}
&\to
\frac{1}{2}
\hat{\bm\theta}^\top
\ofim(\hat{\bm\theta})\ofim(\hat{\bm\theta})^+
\diag{\frac{1}{\bm\sigma}}
\hat{\bm\theta}\\
&\le
\frac{1}{2}
\hat{\bm\theta}^\top
\diag{\frac{1}{\bm\sigma}}
\hat{\bm\theta}.
\end{align*}
By assumption (A5), the RHS is $O(1)$. This term is therefore dropped.
We get
\begin{align*}
    \calO_G
    &=
    - \log{p}(\bm{X}\,\vert\,\hat{\bm\theta})
      + \frac{1}{2}
      \log \left\vert N \ofim(\hat{\bm\theta})
      \diag{\bm\sigma} + \bm{I}
      \right\vert + O(1).
\end{align*}
Note that $\rank{\ofim(\hat{\bm\theta})}\le{D}$, and the matrix $ \ofim(\hat{\bm\theta})\diag{\bm\sigma}$ has the same rank as
$\ofim(\hat{\bm\theta})$.
We can write $\ofim(\hat{\bm\theta})=\bm{L}(\hat{\bm\theta}) \bm{L}(\hat{\bm\theta})^\top$, where $\bm{L}(\hat{\bm\theta})$ has shape $D\times\rank{\ofim(\hat{\bm\theta})}$.
We abuse $\bm{I}$ to denote both the identity matrix of shape $D\times{D}$ and the identity matrix of shape $\rank{\ofim(\hat{\bm\theta})}
\times
\rank{\ofim(\hat{\bm\theta})}$.
By the Weinstein–Aronszajn identity,
\begin{align*}
    \calO_G
    &=
    - \log{p}(\bm{X}\,\vert\,\hat{\bm\theta})
      + \frac{1}{2}
      \log \left\vert N
      \bm{L}(\hat{\bm\theta})
      \bm{L}(\hat{\bm\theta})^\top
      \diag{\bm\sigma} + \bm{I}
      \right\vert + O(1)\\
    &=
    - \log{p}(\bm{X}\,\vert\,\hat{\bm\theta})
      + \frac{1}{2}
      \log \left\vert N
      \bm{L}(\hat{\bm\theta})^\top
      \diag{\bm\sigma}
      \bm{L}(\hat{\bm\theta})
      + \bm{I}
      \right\vert + O(1)\\
     &=
     - \log{p}(\bm{X}\,\vert\,\hat{\bm\theta})
     + \frac{\rank{\ofim(\hat{\bm\theta})}}{2} \log N
      + \frac{1}{2}
      \log \left\vert
      \bm{L}(\hat{\bm\theta})^\top
      \diag{\bm\sigma}
      \bm{L}(\hat{\bm\theta})
      + \frac{1}{N}\bm{I}
      \right\vert + O(1).
\end{align*}

Note
$\bm{L}(\hat{\bm\theta})^\top\diag{\bm\sigma} \bm{L}(\hat{\bm\theta})$ has the same set of non-zero eigenvalues as
$
\bm{L}(\hat{\bm\theta})
\bm{L}(\hat{\bm\theta})^\top\diag{\bm\sigma} = \ofim(\hat{\bm\theta})\diag{\bm\sigma}$ , which we denote as
$\lambda_i^+\left(
\ofim(\hat{\bm\theta})\diag{\bm\sigma}\right)$.
Then,
\begin{align*}
    \calO_G
    =&
    - \log{p}(\bm{X}\,\vert\,\hat{\bm\theta})
     + \frac{\rank{\ofim(\hat{\bm\theta})}}{2} \log N\\
    &+
    \frac{1}{2}
    \sum_{i=1}^{\rank{\ofim(\hat{\bm\theta})}}
    \log \left(
    \lambda_i^+\left(
\ofim(\hat{\bm\theta})\diag{\bm\sigma}\right)
+\frac{1}{N}
\right) + O(1).
\end{align*}

Denote the largest and smallest elements of $\bm\sigma$ as $\sigma_{\max}$ and $\sigma_{\min}$, respectively. Then,
\begin{align*}
\bm{L}(\hat{\bm\theta})^\top
\diag{\bm\sigma}
\bm{L}(\hat{\bm\theta})
\preceq
\sigma_{\max}
\bm{L}(\hat{\bm\theta})^\top
      \bm{L}(\hat{\bm\theta}).
\end{align*}
Hence,
\begin{align*}
\frac{1}{2} \log \left\vert
\bm{L}(\hat{\bm\theta})^\top
\diag{\bm\sigma}
\bm{L}(\hat{\bm\theta})
+
\frac{1}{N}\bm{I}
\right\vert
&\le
    \frac{1}{2}
      \log \left\vert
      \sigma_{\max}
      \bm{L}(\hat{\bm\theta})^\top
      \bm{L}(\hat{\bm\theta})
      + \frac{1}{N}\bm{I}
      \right\vert\\
&=
      \frac{1}{2}
      \sum_{i=1}^{\rank{\ofim(\hat{\bm\theta})}}
      \log
      \left( \sigma_{\max}
      \lambda_i^+(\ofim(\hat{\bm\theta}))
      + \frac{1}{N} \right).
\end{align*}
Similarly,
\begin{align*}
\frac{1}{2} \log \left\vert
\bm{L}(\hat{\bm\theta})^\top
\diag{\bm\sigma}
\bm{L}(\hat{\bm\theta})
+
\frac{1}{N}\bm{I}
\right\vert
\ge
      \frac{1}{2}
      \sum_{i=1}^{\rank{\ofim(\hat{\bm\theta})}}
      \log
      \left(
      \sigma_{\min}
      \lambda_i^+(\ofim(\hat{\bm\theta}))
      + \frac{1}{N}
      \right).
\end{align*}

If $\bm\sigma=\sigma\bm{1}$, then $\sigma_{\max}=\sigma_{\min}=\sigma$.
Both ``$\le$'' and ``$\ge$'' in the above inequalities become tight.

\section{Probability Measures on $\calM$}

Probability measures are not defined on the lightlike $\calM$, because
along the lightlike geodesics, the distance is zero.
To compute the integral of a given function $f(\bm\theta)$ on $\calM$
one has to first choose a proper Riemannian submanifold
$\calM^s\subset\calM$ specified by an embedding $\bm\theta(\bm\theta^s)$,
whose metric is not singular. Then, the integral on $\calM^s$ can be defined as
$\int_{M^s} f\left( \bm\theta(\bm\theta^s) \right) d\bm\theta^s$,
where $\calM^s$ is the sub-manifold associated with the frame $\bm\theta^s=(\theta^1,\cdots,\theta^d)$,
so that $\calT\calM^s=\mathcal{S}(\mathcal{TM})$, and the induced Riemannian volume element as
\begin{align}\label{eq:vol0}
\mathrm{d}\bm\theta^s
&= \sqrt{\vert\fim(\bm\theta^s)\vert}\;d\theta^{1}\wedge{}d\theta^2\wedge\cdots\wedge{}d\theta^d
\nonumber\\ &
= \sqrt{\vert\fim(\bm\theta^s)\vert}\;\deu\bm\theta^s,
\end{align}
where $\deu\bm\theta$ is the Euclidean volume element.
We artificially shift $\mathcal{\bm\theta}$ to be positive definite
and define the volume element as
\begin{align}\label{eq:vol}
\mathrm{d}\bm\theta
&\defeq
\sqrt{\vert\fim(\bm\theta)+\varepsilon_1\bm{I}\vert}\;d\theta^{1}\wedge{}d\theta^2\wedge\cdots\wedge{}d\theta^D
\nonumber\\
& = \sqrt{\vert\fim(\bm\theta)+\varepsilon_1\bm{I}\vert}\;\deu\bm{\theta}^s,
\end{align}
where $\varepsilon_1>0$ is a very small value as compared to the scale of
$\fim(\bm\theta)$ given by $\frac{1}{D}\trace(\fim(\bm\theta))$,
\ie the average of its eigenvalues.
Notice this element will vary with $\bm\theta$: different coordinate systems
will yield different volumes. Therefore it depends on how $\bm\theta$ can be
uniquely specified. This is roughly guaranteed by our assumption that the $\bm\theta$-coordinates correspond to the input coordinates (weights and
biases) up to an orthogonal transformation.  Despite that \cref{eq:vol} is a
loose mathematical definition, it makes intuitive sense and is convenient for
making derivations. Then, we can integrate functions
\begin{equation}\label{eq:int}
\int_{\calM} f(\bm\theta) d\bm\theta
=
\int f(\bm\theta)
\sqrt{\vert\fim(\bm\theta)+\varepsilon_1\bm{I}\vert}\;\deu\bm\theta,
\end{equation}
where the RHS is an integration over $\Re^D$, assuming $\bm\theta$ is real-valued.

Using this tool, we first consider Jeffreys' non-informative prior on
a sub-manifold $\calM^s$ , given by
\begin{equation}
p_{\mathrm{J}}(\bm\theta^s) =
\frac{\sqrt{\vert\fim(\bm\theta^s)\vert}}{\int_{\calM^s}
\sqrt{\vert\fim(\bm\theta^s)\vert}\deu\bm{\theta}^s}.
\end{equation}
It is easy to check $\int_{\calM^s} p(\bm\theta^s) \deu\bm{\theta}^s=1$.
This prior may lead to similar results as \cite{rFIA,bMDL}, \ie a ``razor'' of the model $\calM^s$.
However, we will instead use a Gaussian-like prior, because Jeffreys' prior is not well defined on
$\calM$. Moreover, the integral
$\int_{\calM^s}\sqrt{\vert\fim(\bm\theta^s)\vert}\deu\bm{\theta}^s$
is likely to diverge based on our revised volume element in \cref{eq:vol}.
If the parameter space is real-valued, one can easily check that, the volume based on \cref{eq:vol}
along the lightlike dimensions will diverge.
The zero-centered Gaussian prior corresponds to a better \emph{code}, because it is
commonly acknowledged that one can achieve the same training error and
generalization without using large weights.  For example, regularizing the norm
of the weights is widely used in deep learning. By using such an informative
prior, one can have the same training error in the first term in \cref{eq:mdl},
while having a smaller ``complexity'' in the rest of the terms,
because we only encode such models with constrained weights.
Given the DNN, we define an \emph{informative prior} on the lightlike neuromanifold
\begin{equation}\label{eq:prior}
p(\bm\theta)
=
\frac{1}{V}
\exp\left(-\frac{1}{2\varepsilon_2^2}\Vert\bm\theta\Vert^2\right)
\sqrt{\vert\fim(\bm\theta)+\varepsilon_1\bm{I}\vert},
\end{equation}
where $\varepsilon_2>0$ is a scale parameter of $\bm\theta$,
and $V$ is a normalizing constant to ensure $\int p(\bm\theta)\deu\bm\theta=1$.
Here, the base measure is the Euclidean volume element
$\deu\bm\theta$, as
$\sqrt{\vert\fim(\bm\theta)+\varepsilon_1\bm{I}\vert}$ already appeared in
$p(\bm\theta)$.
Keep in mind, again, that this $p(\bm\theta)$ is defined in a
special coordinate system, and is not invariant to re-parametrization.
This distribution is also isotropic in the
input coordinate system, which agrees with initialization techniques\footnote{Different layers, or weights and biases, may use different variance
in their initialization. This minor issue can be solved by a simple re-scaling
re-parameterization.}.

This bi-parametric prior connects Jeffreys' prior (that is widely used in MDL) and a Gaussian prior (that is widely used in deep learning).
If $\varepsilon_2\to\infty$, $\varepsilon_1\to0$, it coincides with Jeffreys'
prior (if it is well defined and $\fim(\bm{\theta})$ has full rank); if $\varepsilon_1$ is large,
the metric $\left(\fim(\bm\theta)+\varepsilon_1\bm{I}\right)$ becomes spherical,
and \cref{eq:prior} becomes a Gaussian prior.
We refer the reader to \cite{alphaprior-2005,weylprior} for other extensions of Jeffreys' prior.

The normalizing constant of \cref{eq:prior} is an information volume measure of $\calM$, given by
\begin{align}
V \defeq \int_{\calM} \exp\left(-\frac{1}{2\varepsilon_2^2} \Vert\bm\theta\Vert^2 \right) \mathrm{d}\bm{\theta}.
\end{align}
Unlike Jeffreys' prior whose information volume (the 3rd term on the RHS of
\cref{eq:mdl}) can be unbounded, this volume can be bounded as stated in the following theorem.
\begin{theorem}\label{thm:bound}
\begin{equation}
(\sqrt{2\pi\varepsilon_1}\varepsilon_2)^{D}
\le{}V\le{}
(\sqrt{2\pi(\varepsilon_1+\lambda_m)}\varepsilon_2)^{D},
\end{equation}
where $\lambda_m$ is the largest eigenvalue of the FIM $\fim(\bm\theta)$.
\end{theorem}
Notice $\lambda_m$ may not exist, as the integration is taken over
$\bm\theta\in\calM$.
Intuitively, $V$ is a weighted volume \wrt a Gaussian-like prior distribution on
$\calM$, while the 3rd term on the RHS of \cref{eq:mdl} is an unweighted
volume.
The larger the radius $\varepsilon_2$, the more ``number'' or
possibilities of DNNs are included; the larger the parameter $\varepsilon_1$, the
larger the local volume element in \cref{eq:vol} is measured, and therefore
the total volume is measured larger.
$\log{V}$ is an $O(D)$ terms, meaning the volume grows with the number of dimensions.

\subsection{Proof of \Cref{thm:bound}}

By definition,
\begin{align*}
V =
\int_{\calM} \exp\left(-\frac{1}{2\varepsilon_2^2} \Vert\bm\theta\Vert^2 \right) \mathrm{d}\bm{\theta}
=
\int \exp\left(-\frac{1}{2\varepsilon_2^2}\Vert\bm\theta\Vert^2\right)
\sqrt{\vert\fim(\bm\theta)+\varepsilon_1\bm{I}\vert}
\deu\bm\theta.
\end{align*}
By our assumption, $\bm\theta$ is an orthogonal transformation of the neural network weights
and biases, and therefore $\bm\theta\in\Re^D$.
We have
\begin{align*}
\sqrt{\vert\fim(\bm\theta)+\varepsilon_1\bm{I}\vert}
\ge
\sqrt{\vert\varepsilon_1\bm{I}\vert}
=
\varepsilon_1^{\frac{D}{2}}.
\end{align*}
Hence
\begin{align*}
V
&\ge
\int \exp\left(-\frac{1}{2\varepsilon_2^2}\Vert\bm\theta\Vert^2\right)
\varepsilon_1^{\frac{D}{2}}
\deu\bm\theta\nonumber\\
&=
(2\pi)^{\frac{D}{2}}
\varepsilon_2^D
\varepsilon_1^{\frac{D}{2}}
\int \exp\left(
-\frac{D}{2}\log2\pi
-\frac{1}{2}\log\vert\varepsilon_2^2\bm{I}\vert
-\frac{1}{2\varepsilon_2^2}\Vert\bm\theta\Vert^2\right)
\deu\bm\theta\nonumber\\
&=
(2\pi)^{\frac{D}{2}}
\varepsilon_2^D
\varepsilon_1^{\frac{D}{2}}
=\left(\sqrt{2\pi\varepsilon_1}\varepsilon_2\right)^{D}.
\end{align*}
For the upper bound, we prove a stronger result as follows.
\begin{equation*}
\sqrt{\vert\fim(\bm\theta)+\varepsilon_1\bm{I}\vert}
=
\left(
\prod_{i=1}^D \left(\lambda_i+\varepsilon_1\right)^{\frac{1}{D}}
\right)^{\frac{D}{2}}
\le
\left(
\frac{1}{D} \trace(\fim(\bm\theta)) + \varepsilon_1
\right)^{\frac{D}{2}}.
\end{equation*}
Therefore
\begin{equation*}
V \le
\left(\sqrt{2\pi}\varepsilon_2\right)^{D}
\left(
\frac{1}{D} \trace(\fim(\bm\theta)) + \varepsilon_1
\right)^{\frac{D}{2}}.
\end{equation*}
If one applies
$\frac{1}{D} \trace(\fim(\bm\theta)) \le \lambda_m$ to the RHS, the
upper bound is further relaxed as
\begin{equation*}
V \le
\left(\sqrt{2\pi}\varepsilon_2\right)^{D}
\left( \lambda_m + \varepsilon_1 \right)^{\frac{D}{2}}
=
\left(\sqrt{2\pi (\varepsilon_1+\lambda_m)}\varepsilon_2\right)^{D}.
\end{equation*}

\section{An Alternative Derivation of the Razor}

In this section, we provide an alternative derivation of the propose razor
$\calO$ based on a different prior. The main observations on the negative
complexity are consistent with the cases of Gaussian and Jeffreys' priors.

We plug in the expression of $p(\bm\theta)$ in \cref{eq:prior} and get
\begin{align*}
-\log p(\bm{X}) \approx \;
&
-\log{p}(\bm{X}\,\vert\,\hat{\bm\theta}) + \log{V}
\nonumber\\
&-\log\int_{\calM} \left( -\frac{\Vert\bm\theta\Vert^2}{2\varepsilon_2^2}
-\frac{N}{2} (\bm\theta-\hat{\bm\theta})^\top \ofim(\hat{\bm\theta}) (\bm\theta-\hat{\bm\theta}) \right) d\bm\theta.
\end{align*}
In the last term on the RHS, inside the parentheses is a quadratic function \wrt $\bm\theta$.
However the integration is \wrt to the non-Euclidean volume element $d\bm\theta$
and therefore does not have closed form. We need to assume
\begin{description}
\item[(A7)] $N$ is large enough so that
$\vert\fim(\bm\theta)+\varepsilon_1\bm{I}\vert\approx
\vert\fim(\hat{\bm\theta})+\varepsilon_1\bm{I}\vert$.
\end{description}
This means the quadratic function will be sharp enough to make the volume element
$\mathrm{d}\bm\theta$ to be roughly constant.
Along the lightlike dimensions (zero eigenvalues of $\fim(\bm\theta)$) this is trivial.

Plug \cref{eq:prior} into \cref{eq:px}, the following three terms
\begin{equation*}
\frac{1}{V},\quad
\sqrt{\vert\fim(\bm\theta)+\varepsilon_1\bm{I}\vert}
\approx
\sqrt{\vert\fim(\hat{\bm\theta})+\varepsilon_1\bm{I}\vert},
\quad
\exp\left( \log{p}(\bm{X}\,\vert\,\hat{\bm\theta}) \right)
= p(\bm{X}\,\vert\,\hat{\bm\theta})
\end{equation*}
can all be taken out of the integration as constant scalers, as they do not depend on $\bm\theta$.
The main difficulty is to perform the integration
\begin{align*}
&\int \exp\left( -\frac{\Vert\bm\theta\Vert^2}{2\varepsilon_2^2}
-\frac{N}{2} (\bm\theta-\hat{\bm\theta})^\top \ofim(\hat{\bm\theta}) (\bm\theta-\hat{\bm\theta}) \right) \deu\bm\theta\\
=&
\int \exp\left(
-\frac{1}{2}\bm\theta^\top \bm{A}\bm\theta
+ \bm{b}^\top \bm\theta
+ c
\right) \deu\bm\theta\\
=&
\int\exp\left(-\frac{1}{2}(\bm\theta-\bm{A}^{-1}\bm{b})^\top\bm{A}
(\bm\theta-\bm{A}^{-1}\bm{b})
+
\frac{1}{2}\bm{b}^\top{A}^{-1}\bm{b} +c
\right) \deu\bm\theta\\
=&
\exp\left( \frac{1}{2}\bm{b}^\top{A}^{-1}\bm{b} +c \right)
\int\exp\left(-\frac{1}{2}(\bm\theta-\bm{A}^{-1}\bm{b})^\top\bm{A}
(\bm\theta-\bm{A}^{-1}\bm{b})\right)
\deu\bm\theta\\
=&
\exp\left( \frac{1}{2}\bm{b}^\top{A}^{-1}\bm{b} +c \right)
\exp\left(\frac{D}{2}\log2\pi - \frac{1}{2}\log\vert\bm{A}\vert
\right)\\
=&
\exp\left( \frac{1}{2}\bm{b}^\top{A}^{-1}\bm{b} +c
+ \frac{D}{2}\log2\pi - \frac{1}{2}\log\vert\bm{A}\vert
\right).
\end{align*}
where
\begin{equation*}
\bm{A} = N\ofim(\hat{\bm\theta}) + \frac{1}{\varepsilon_2^2}\bm{I},\quad
\bm{b} = N\ofim(\hat{\bm\theta}) \hat{\bm\theta},\quad
c =-\frac{1}{2} \hat{\bm\theta}^\top N\ofim(\hat{\bm\theta}) \hat{\bm\theta}.
\end{equation*}
The rest of the derivations are straightforward.
Note $R= -c - \frac{1}{2}\bm{b}^\top{A}^{-1}\bm{b}$.

After derivations and simplifications, we get
\begin{align}
&-\log p(\bm{X}) \approx
 - \log{p}(\bm{X}\,\vert\,\hat{\bm\theta}) + \frac{D}{2}\log\frac{N}{2\pi} + \log{V}
\nonumber\\
& + \frac{1}{2}\log\left\vert \ofim(\hat{\bm\theta}) + \frac{1}{N\varepsilon_2^2}\bm{I}\right\vert
 -\frac{1}{2}\log \left\vert\fim(\hat{\bm\theta})+\varepsilon_1\bm{I}\right\vert
+ R.
\end{align}
The remainder term is given by
\begin{align}\label{eq:remainder}
R
= & \frac{1}{2}{\hat{\bm\theta}^\top} \bigg[ N\ofim(\hat{\bm\theta})
- N\ofim(\hat{\bm\theta}) \bigg(N\ofim(\hat{\bm\theta})+\frac{1}{\varepsilon_2^2}\bm{I}\bigg)^{-1} N\ofim(\hat{\bm\theta})
\bigg]
\hat{\bm\theta}.
\end{align}
We need to analyze the order of this $R$ term.
Assume the largest eigenvalue of $\ofim(\hat{\bm\theta})$ is $\lambda_m$, then
\begin{align}
\vert{R}\vert \le
\frac{N\lambda_m}{\varepsilon_2^2N\lambda_m+1}
\Vert\hat{\bm\theta}\Vert^2.
\end{align}
We assume
\begin{description}
\item[(A8)]
The ratio between the scale of each dimension of the MLE $\hat{\bm\theta}$
to $\varepsilon_2$,
\ie $\frac{\hat{\theta}_i}{\varepsilon_2}$ ($i=1,\cdots,D$)
 is in the order $O(1)$.
\end{description}
Intuitively, the scale parameter $\varepsilon_2$ in our prior $p(\bm\theta)$ in
\cref{eq:prior} is chosen to ``cover'' the good models.  Therefore, the order of
$R$ is $O(D)$.
As $N$ turns large, $R$ will be dominated by the 2nd $O(D\log{N})$ term.
We will therefore discard $R$ for simplicity. It could be useful for a more
delicate analysis. In conclusion,
we arrive at the following expression
\begin{align}\label{eq:razor}
\calO
 \defeq
& -\log{p}(\bm{X}\,\vert\,\hat{\bm\theta}) + \frac{D}{2}\log\frac{N}{2\pi} + \log{V}
+\frac{1}{2} \log \frac{\left\vert \ofim(\hat{\bm\theta}) + \frac{1}{N\varepsilon_2^2}\bm{I}\right\vert}
{\left\vert\fim(\hat{\bm\theta})+\varepsilon_1\bm{I}\right\vert}.
\end{align}

Notice the similarity with \cref{eq:mdl},
where the first two terms on the RHS are exactly the same.
The 3rd term is an $O(D)$ term, similar to the 3rd term in \cref{eq:mdl}.
It is bounded according to \cref{thm:bound}, while the 3rd term in \cref{eq:mdl} could be unbounded.
Our last term is in a similar form to the last term in \cref{eq:mdl}, except it
is well defined on lightlike manifold.  If we let $\varepsilon_2\to\infty$,
$\varepsilon_1\to0$, we get exactly \cref{eq:mdl} and in this case
$\calO=\chi$.  As the number of parameters $D$ turns large, both the 2nd
and 3rd terms will grow linearly \wrt $D$, meaning that they contribute
positively to the model complexity.
Interestingly, the fourth term is a ``\emph{negative complexity}''.
Regard $\frac{1}{N\varepsilon_2^2}$ and $\epsilon_1$ as small positive values.
The fourth term essentially is a $\log$-ratio from the observed FIM to the true
FIM. For small models, they coincide, because the sample size $N$ is large
based on the model size. In this case, the effect of this term is minor.
For DNNs, the sample size $N$ is very limited based on the huge model size $D$.
Along a dimension $\theta_i$, $\ofim(\bm\theta)$ is likely to be singular
as stated in \cref{thm:dim}, even if $\fim$ has a very small positive value.
In this case, their log-ratio will be negative.
Therefore, the razor $\calO$ favors DNNs with their Fisher-spectrum
clustered around 0.

\begin{figure}[tb]
\centering\includegraphics[width=.4\textwidth]{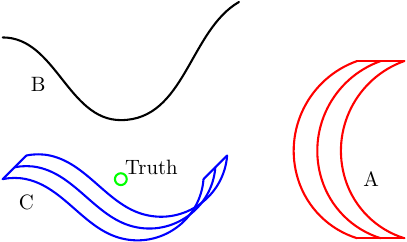}
\caption{A: a model far from the truth (underlying distribution of observed data);
B: close to the truth but sensitive to parameter;
C (deep learning): close to the truth with many good local optima.\label{fig:models}}
\end{figure}

In \cref{fig:models}, model C displays the concepts of a DNN, where there are many good local optima.
The performance is not sensitive to specific values of model parameters.
On the lightlike neuromanifold
$\calM$, there are many directions that are very close to being lightlike.
When a DNN model varies along these directions, the model slightly changes in
terms of $\fim(\bm\theta)$, but their prediction on the samples measured
by $\ofim(\bm\theta)$ are invariant.  These directions count
\emph{negatively} towards the complexity, because these extra freedoms
(dimensions of $\bm\theta$) occupy almost zero volume in the geometric sense,
and are helpful to give a shorter code to future unseen samples.

To obtain a simpler expression,
we consider the case that $\fim(\bm\theta)\equiv\fim(\hat{\bm\theta})$ is both constant
and diagonal in the interested region defined by \cref{eq:prior}. In this case,
\begin{equation}
\log{V} \approx \frac{D}{2}\log2\pi + D\log{}\varepsilon_2 + \frac{1}{2}\log\vert\fim(\hat{\bm\theta})+\varepsilon_1\bm{I}\vert.
\end{equation}
On the other hand, as $D\to\infty$, the spectrum of the FIM $\fim(\bm\theta)$
will follow the density $\rho_{\fim}(\bm\theta)$.
We plug these expressions into \cref{eq:razor}, discard all lower-order terms, and get
a simplified version of the razor
\begin{align}\label{eq:razorsimple}
\calO \approx
& - \log p(\bm{X} \,\vert\, \hat{\bm\theta}) + \frac{D}{2}\log{N}
+ \frac{D}{2}\int_{0}^{\infty} \rho_{\fim}(\lambda) \log\left( \lambda + \frac{1}{N\varepsilon_2^2} \right) \mathrm{d}\lambda,
\end{align}
where $\rho_{\fim}$ denotes the spectral density of the Fisher information matrix.

 \end{appendices}

\end{document}